\documentclass[journal]{IEEEtran}
\usepackage{amsmath,graphicx}
\usepackage{epsfig}
\usepackage{multirow,bigstrut}
\usepackage{amssymb}
\usepackage{color}
\usepackage{caption}
\usepackage{url}
\usepackage{booktabs}
\usepackage{diagbox}
\usepackage{subcaption}
\usepackage[numbers]{natbib}
\newlength\savewidth

\newcommand{\M}[1]{{\color{black}#1}}
\newcommand{\YY}[1]{{\color{black}#1}}
\newcommand{\md}[1]{{\color{black}#1}}
\newcommand{\RR}[1]{{\color{black}#1}}
\newcommand{\RRR}[1]{{\color{black}#1}}

\begin{document}
%

\title{\M{Progressive Spatial Recurrent Neural Network \\for Intra Prediction}}


%
%
%

\author{
 Yueyu Hu, Wenhan Yang,~\IEEEmembership{Member,~IEEE,} Mading Li, Jiaying Liu,~\IEEEmembership{Senior Member,~IEEE}\thanks{ This work was supported in part by National Natural Science Foundation of China under contract No. 61772043, and in part by Beijing Natural Science Foundation under contract No. L182002 and No. 4192025. (Corresponding author: Jiaying Liu.)} \thanks{ \RR{Yueyu Hu, Wenhan Yang, Mading Li, and Jiaying Liu are with Institute of Computer Science and Technology, Peking University, Beijing, 100080, China, e-mail: \{huyy, yangwenhan, martinli0822, liujiaying\}@pku.edu.cn.} }}

%
%

\markboth{IEEE Transactions on Multimedia}%
{Shell \MakeLowercase{\textit{et al.}}: Bare Demo of IEEEtran.cls for IEEE Journals}
%



\maketitle

\begin{abstract}

Intra prediction is an important component of modern video codecs, which is able to efficiently squeeze out the spatial redundancy in video frames.
\md{With preceding pixels as \YY{the} context, traditional intra prediction schemes generate linear predictions based on several predefined directions (\textit{i.e.} modes) for blocks to be encoded. However, these modes are relatively simple and their predictions may fail when facing blocks with complex textures, which leads to additional bits encoding the residue. In this paper, we design a Progressive Spatial Recurrent Neural Network (PS-RNN) that learns to conduct intra prediction. Specifically, our PS-RNN consists of three spatial recurrent units and progressively generates predictions by passing information along from preceding contents to blocks to be encoded.
To make our network generate predictions considering both distortion and bit-rate, we propose to use Sum of Absolute Transformed Difference (SATD) as \YY{the} loss function to train PS-RNN since SATD is able to measure rate-distortion cost of encoding a residue block.
\YY{Moreover,
our method supports variable-block-size for intra prediction, which is more practical in real coding conditions.}}
The proposed intra prediction scheme achieves on average 2.5\% bit-rate reduction on variable-block-size settings under the same reconstruction quality compared with HEVC.

\end{abstract}


%
\IEEEpeerreviewmaketitle

\section{Introduction}
%
%
%
%
\IEEEPARstart{I}NTRA prediction efficiently \M{reduces} \md{spatial redundancy in videos} and improves video coding performance.
It has been adopted in modern \md{codecs} like H.264/AVC \cite{wiegand2003overview} and HEVC~\cite{lainema2012intra}. Compared with H.264/AVC, HEVC achieves a leap in rate-distortion performance by enriching reference samples and enlarging the number of angular modes \md{in intra prediction}. Besides, HEVC adopts a more flexible quadtree coding structure, which adaptively chooses the appropriate block size during the coding process.
\md{
However, there are some drawbacks in traditional intra prediction schemes. On one hand, modern codecs only use a single line of preceding reconstructed pixels above and on the left side of the \YY{current prediction unit (PU) as the reference to generate predictions, which can be affected by the noise (\textit{e.g.} quantization noise) in the reconstructed pixels.} On the other hand,} for directional intra prediction in HEVC, only Planar, DC and other 33 angular modes are utilized. None of above can handle complex texture even with Rate-Distortion Optimization (RDO)~\cite{sullivan1998rate}.

To address the drawbacks of the single-line reference scheme, improved intra prediction with multi-line reference scheme~\cite{li2016efficient,li2017intra} or using filtered reference samples~\cite{wei2016improved} are investigated. \YY{Furthermore}, to enhance the ability of HEVC intra prediction for complex textures, synthesis-based methods~\cite{yeh2015predictive}, copying-based methods~\cite{chen2016improving,zhang2015hybrid} and inpainting-based methods~\cite{qi2012intra,liu2007image} are developed. Although these methods \M{mitigate} the problems, \M{each of them} has limitations. Though some methods adopt the multi-line reference scheme, the contextual information of \YY{reference pixels} is usually not utilized. \YY{Besides, the performance of copying-based methods is limited by the structural similarities of intra patches, while inpainting-based methods are not capable of accurately predicting the pixels at the bottom-right of the current PU.}

\RRR{
In the past decades, Deep Neural Network (DNN), as an effective data-driven solution for computer vision tasks, has
been exploited to accomplish high-level visual recognition tasks, \textit{e.g.} image classification \cite{he2016deep}, action recognition \cite{li2018bilinear,song2018spatio}, and low-level image restoration/enhancement tasks, including super-resolution \cite{liu2016retrieval,yang2017deep,yang2018reference}, low-light enhancement \cite{wei2018light,wang2018gladnet}, rain removal \cite{yang2019joint,liu2019d3r,yang2019scale,yang2019jorder2}, \textit{etc}.
With powerful computing devices like GPUs and TPUs~\cite{jouppi2017datacenter}, given well-defined inputs and outputs, the network can automatically learn an end-to-end mapping from inputs to outputs. This technique has been utilized in video coding tasks, \textit{e.g.}, fractional interpolation~\cite{xia2018dcc}, inter prediction~\cite{liu2019one,xia2019iter}, loop filter~\cite{zhang2018dmcnn,wang2019loop}, and fast mode decision~\cite{liu2016cnn}. 

Deep learning can also facilitate intra prediction. Pioneering works that utilize deep learning models in intra prediction explore the potential of Fully-Connected (FC) neural networks \cite{li2017icip,li2018tip}. With end-to-end learning and the block-level reference scheme, the novel predictor can utilize more information in the context and largely reduce quantization noise. It achieves a great leap in rate-distortion performance over original HEVC standard. Recently in the work of \cite{dumas2018context}, by combining FC networks and CNNs, the prediction accuracy is further enhanced and the efficiency of the predictor is improved, which makes the technique of learned intra prediction practical for video and image compression. However, there is still much room for improvement. Most of FC and CNN networks take the symmetric inputs and make use of information of the neighboring pixels in an isotropic way, which ignores the directional spatial redundancy of image modeling. Furthermore, FC networks include a large amount of parameters. Without explicit constraints and well-designed network structures, it is hard to obtain a good FC model.
}


\RRR{
Considering these limitations, more advanced models are called to provide more flexible and enhanced modeling capacity. Compared to CNN and FC, the emerging RNNs improve the way of information utilization and image modeling.
It can be regarded as the extension of CNN, which has an asymmetric kernel and enables the information propagation and aggregation in an anisotropic way. Taking asymmetric inputs, RNNs facilitate better directional and structural modeling. Therefore, it is a trend recently to utilize spatial RNNs for image restoration and generation tasks \cite{liu2017learning,liu2016learning,van2016pixel}. By unrolling the 2D images and updating the information in the spatial domain progressively, RNNs better model pixel-level and feature-level dependencies in images. This capability facilitates local information aggregation and enhances the overall modeling capacity. Therefore, RNN is a desirable framework for intra-prediction whose input is non-symmetric. That is, for a map of pixels or features, the reference information is located on the left and top of the map, while a large area on the right bottom of the map contains no useful information.

In this paper, we aim to address the drawbacks of traditional intra prediction schemes and aforementioned deep learning based methods. Specifically, we build a Progressive Spatial Recurrent Neural Network (PS-RNN) based on spatial RNNs to learn to predict contents of PUs in intra prediction.} Three spatial recurrent units are stacked sequentially, which update and aggregate internal memory progressively along certain directions (\textit{i.e.} horizontal and vertical in our work), facilitating the modeling capacity of PS-RNN. To further improve the coding performance of our method, we propose to use \YY{SATD} as the training loss of PS-RNN. SATD not only calculates the distortion but also reflects bit-rate, thus it is a good criterion to guide the network training for intra prediction. \RR{Moreover, our method enables variable-block-size configuration, which further reduces bit-rate and makes the method practical for video coding applications. The method is implemented and evaluated on HEVC as the anchor. Note that as the proposed method is a progressive model, it can also be applied to other hybrid video or image codecs like Versatile Video Coding (VVC)~\cite{vvc}, AOMedia Video 1 (AV1)~\cite{chen2018overview} and H.264/AVC~\cite{wiegand2003overview}. For non-square partitioning conditions in these codecs, the model can be set up to progressively generate the prediction signal line-by-line. Beyond that, the proposed method is a general method and can be adopted by a video or image codec if the following two properties are satisfied. First, intra prediction tools are adopted in the codec. Second, the prediction is conducted in the pixel domain. The above-mentioned properties are almost satisfied for all modern hybrid codecs.}

To summarize, our contributions are listed as follows:
\begin{itemize}
\md{
\RRR{
\item We propose a progressive neural network that embeds spatial RNNs for intra prediction. The network takes the advantages of RNNs and block-wise reference scheme. The information from the reference blocks is explicitly propagated to the to-be-predicted area. Experimental results demonstrate that this deep learning method achieves superior performance compared with methods using one single line as the reference, especially when faced with severe quantization noises in low-bit-rate configurations and complex textures.}
\item A progressive spatial RNN is specially designed for intra prediction. It infers information from reference inputs to the current PU progressively along certain directions resulting in consistent and accurate predictions.
\item \RRR{In order to make our network generate predictions with less distortion and lower bit-rate, we propose to use SATD as the training loss function. Compared to MSE, using SATD as loss function largely improves the rate-distortion performance of the codec. We provide comprehensive experimental evaluations and empirical studies to analyze SATD loss function in network training for video coding.}

\item Our model supports the variable-block-size configuration, which makes our method practical for video coding applications. Allowing variable-block-size coding significantly reduces bit-rate especially for high-resolution videos.
}
\end{itemize}

The rest of the paper is organized as follows. In Section \ref{related-work}, we review related works on deep learning for intra prediction for video coding and deep learning for image compression. Section \ref{proposed-method} introduces our proposed intra prediction. We formulate the problem and analyze the proposed PS-RNN trained with SATD loss function in detail. In Section \ref{experimental-results}, we show our gain in rate-distortion performance compared with HEVC and previous methods. We qualitatively and quantitatively compare the predictions of our method and other methods to explain where the improvement comes from. Finally in Section \ref{conclusion}, we draw a conclusion of this paper.

\section{Related Work} \label{related-work}

\subsection{Intra Prediction for Video Coding}


Modern codecs usually consist of multiple parts to progressively squeeze out redundancy and reduce the bit-rate \cite{sullivan2012overview,wiegand2003overview}. In this work, we focus on the intra prediction component of video coding methods. In HEVC, 35 intra prediction modes are assembled with RDO to predict for the encoding block, including Planar mode and DC mode. DC mode fills the block with DC signals. If Planar mode is selected, each pixel of the block is generated by a linear combination of corresponding pixels in the reference samples. For the other 33 modes, the predition signals are generated according to reconstructed pixels on predefined directions. 
\RR{
However, this reference scheme sometimes fails in some hard cases, especially for low-bit-rate coding conditions and frames with complex textures. In low-bit-rate conditions, large quantization step size results in heavy quantization noise and interferes the prediction. To overcome this vulnerability to noise, intra prediction involving multiple reference lines is developed to jointly take additional reference lines as reference samples~\cite{li2016efficient,li2017intra}. It reduces the impact of the noise to some extent.

Tackling the inability to handle complex texture is more challenging, as the inability lies in the design of angular prediction modes. Emerging end-to-end methods based on deep neural networks have been initially studied in recent years.} Deep neural networks can automatically learn the end-to-end mapping of inputs and outputs. It can also be easily accelerated using large-scale parallel programming. In \cite{liu2016cnn}, CNN has been utilized for mode decision as it has a strong potential of capturing global feature from image data.  In \cite{li2017icip,cui2017convolutional}, FC network and CNN are exploited directly in intra prediction. By training a network to build a mapping from the reference samples to the prediction signal, FC networks and CNN show improvement in rate-distortion performance compared with HEVC.
\begin{figure*}[htb]
    \centering
    \includegraphics[width=0.9\linewidth]{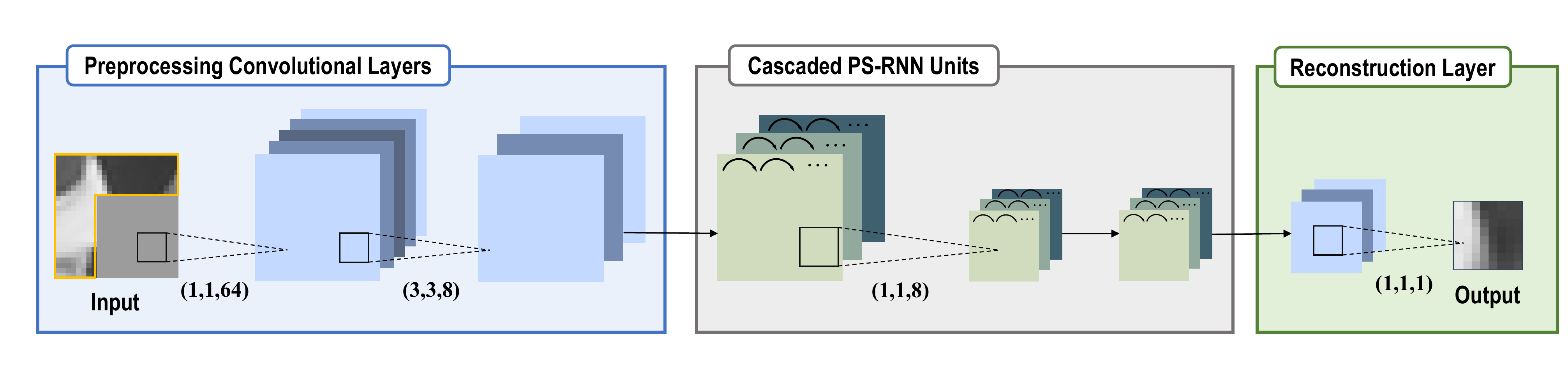}
    \caption{Architecture of PS-RNN. Preprocessing convolutional layers map the input image into feature space. After the preprocessing, the feature maps are handled by cascaded PS-RNN units. A spatial down-sampling is performed on the output of the first PS-RNN unit, making the feature maps identical to the PU in spatial scale. A convolutional reconstruction layer maps the predicted feature maps to pixel space.}
    \label{fig:arch}
\end{figure*}
There are two main problems for the deep-network-based intra prediction methods. On one hand, the network is usually trained using pixel-wise MSE loss function. But the final coding cost is also influenced by the correlations between adjacent pixels. Thus, two predictions with similar MSE can have a much different coding cost. On the other hand, CNN is not capable of handling asymmetric image completion tasks, as the whole input block is convolved unconditionally. Large areas with no texture information interferes the extraction of spatial features. To address these issues, we proposed the spatial RNN for intra prediction with SATD loss function. The network progressively handles prediction contexts, mitigating the influence of asymmetric distribution of reference pixels. Besides, SATD is utilized as the loss function for training. In HEVC, SATD is widely used to measure the rate-distortion cost of encoding a residue block. Since it applies a time-frequency transformation to the difference, it can jointly measure the pixel-wise difference and the contextual difference. As a result, SATD is more consistent with the real coding cost than MSE.

\subsection{Deep Learning for Image Compression}
Compression of videos and images shares the same philosophy as videos can be regarded as stacks of images. HEVC also supports encoding images in the still image profile \cite{nguyen2015objective}. Image codecs are relatively simpler than video codecs. So deep learning is first explored for image compression and then extended to videos. There have been several approaches to utilize deep neural networks for lossy image compression. These approaches are mostly based on autoencoders \cite{theis2017lossy,balle2016end,gregor2016towards,agustsson2017soft} and convolutional recurrent neural networks \cite{toderici2017full,baig2017learning}. These methods achieve impressive visual quality with low coding bit-rate and the performance is superior to conventional image codecs like JPEG and JPEG2000. However, these approaches still face challenges for high bit-rate conditions. Besides, even in low bit-rate conditions, they are not yet comparable with intra coding in HEVC still image profile. These methods for image compression are still far from being adopted in video codecs. Different from the methods mentioned above, in our work, we focus on optimizing the intra prediction scheme in the loop of video coding.

\subsection{Image Inpainting for Intra Prediction}
Intra prediction shares similar characteristics with image inpainting, which aims to accurately fill in the missing areas of an image. Previous works on image inpainting study the methods to propagate information from neighboring known pixels to missing areas or to copy similar patches to unknown regions~\cite{ballester2001filling,bertalmio2000image,kwatra2005texture,shuai2018inpaint}. Recently, works studying learning based image inpainting emerge. These methods typically utilize CNN to directly map input images with unknown regions to restored output image \cite{yang2017high,pathak2016context}.

Image inpainting techniques have been employed in image and video coding \cite{liu2007image,baig2017learning,qi2012intra}. However, unlike the original inpainting task where the unkown regions are relatively small and usually surrounded by known pixels, in intra predition, only the preceding pixels are available. Predicting the pixels on the right below of the current PU is hard as this area is too far from the reference. Therefore, inpainting based methods face challenges in intra prediction.

Different from previous works on image inpainting methods, we take the unique structure of intra prediction context into consideration. The predictions are generated in a progressive way. \YY{Thus, the prediction} for the hard regions can take previously filled regions as the reference.

\begin{figure*}[tb]
    \centering
    \includegraphics[width=0.9\linewidth]{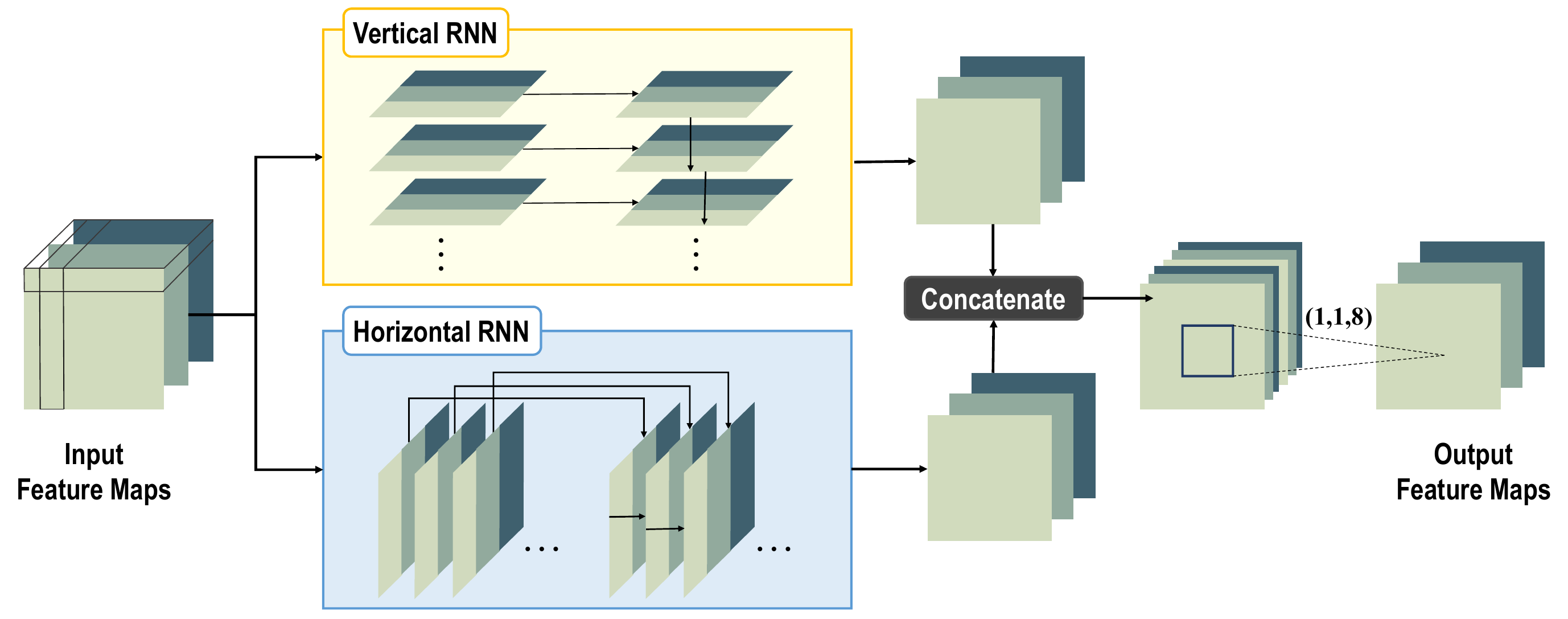}
    \caption{Structure of a PS-RNN unit. It splits a stack of feature maps into vertical and horizontal planes. Each plane represents a feature map of a vertical line or a horizontal line in the original grey-scale image. After the progressive prediction, these planes are concatenated to reconstruct the feature maps. A convolutional layer is used to fuse the predictions from the vertical and horizontal feature maps.}
    \label{fig:unit}
\end{figure*}
\section{PS-RNN for Intra Prediction}\label{proposed-method}

In this section, we present a detailed description and analysis of the proposed PS-RNN. We first formulate the problem, where we conduct the prediction in feature space. Then we illustrate the overall framework of the network. After that we investigate the SATD loss function for training a network for intra prediction. At last, we show how the proposed network is integrated with HEVC with variable-block-size configuration.
\RR{
\subsection{Spatial RNN}
 RNNs are originally designed to process time series data, such as speech, texts and videos. The behavior of RNNs is determined by its basic cell type in the recurrent modeling. Gated Recurrent Units (GRU) \cite{chung2014empirical} and Long Short-Term Memory (LSTM) \cite{hochreiter1997long} are two commonly used RNN structures. They aggregate the information in 1D temporal domain from previous time-steps, update the internal state of each cell given the context of the current time-step. However, they face two gaps when being applied to image processing tasks:
\begin{itemize}
	\item They need to adapt to aggregating information in 2D spatial domain.
	\item They need to model the highly complex mapping of intra-prediction, which might be more challenging than context aggregation in time series modeling.
\end{itemize}
For that, we build our progressive spatial RNN (PS-RNN), with two distinguished properties:
\begin{itemize}
	\item We realize 2-D spatial modeling by stacking spatial RNNs in two orthogonal directions.
	\item We improve the modeling capacity to capture long-term and complex spatial dependency by stacking hierarchical spatial RNNs.
\end{itemize}
With these two considerations, our PS-RNN is better at handling intra-prediction task.
}

\RR{
\RRR{
\subsection{Overall Framework}
\textbf{Motivation}. As is analyzed, there are drawbacks for previously proposed CNNs and FCs, which either fail to fully utilize spatial correlations, or do not well handle the asymmetric shape of context. Recently, RNNs have shown great potential in image and video restoration tasks \cite{liu2017learning,liu2016learning,van2016pixel}. 
With the progressiveness property, RNNs can jointly tackle the above mentioned problems. As it enables the information to propagate in a certain direction, it is also a desirable framework for intra-prediction when we utilize it to propagate information from non-missing regions to missing ones.

\textbf{Overall Model}. Our goal is to accurately generate the prediction signals for the to-be-predicted block given the existing pixels on the left and top. The goal is achieved in three stages, shown in Fig. \ref{fig:arch}. In the first stage, convolutional layers extract local features of the input context block and transform the image to feature space. As the pixels are filtered and abstracted to features, the network is able to reduce quantization noise in the reference pixels. In the second stage, we exploit cascaded PS-RNN units to generate the prediction of the feature vectors. At last, two convolutional layers map the predicted feature vectors back to pixels, which finally form the prediction signals. Though convolutional layers are employed in the framework, these layers are not supposed to directly generate the prediction signals. The size of the kernels in these convolutional layers is small and these layers are designed to extract local features. Thus, they are not affected by the asymmetry of the inputs, which disturbs the reconstruction process of CNN based models. The progressiveness of the PS-RNN unit helps mitigate the problem of the asymmetry. The network is capable of handling intra prediction for a wide variety of contexts in a codec.}

\textbf{PS-RNN Units}. The detailed structure of the PS-RNN units is illustrated in Fig. \ref{fig:unit}. In each unit, the input feature tensor is split to  horizontal and vertical lines, respectively. Suppose the shape of the feature tensor is $(n,n,c)$, with $c$ to be the number of channels. It is split to $\mathbf{X}^h = \{\mathbf{X}_{0 \cdot \cdot}, \mathbf{X}_{1 \cdot \cdot}, \dots, \mathbf{X}_{n-1 \cdot \cdot}\}$ and $\mathbf{X}^v = \{\mathbf{X}_{\cdot 0 \cdot}, \mathbf{X}_{\cdot 1 \cdot}, \dots, \mathbf{X}_{\cdot n-1 \cdot}\}$. Each element in $\mathbf{X}^h$ or $\mathbf{X}^v$ is a feature map of shape $(n,c)$. For simplification of the notations and without loss of generality, we take the horizontal form as an example. To conduct recurrent learning, each element of shape $(n,c)$ in the sequence is flattened to a vector of $n\times c$ dimensions. We define the $t$-th feature vector in the sequence as $\mathbf{x}_t$. The defination of $\mathbf{X}^h$ can be simplified as $\mathbf{X}^h = \{\mathbf{x}_0, \mathbf{x}_1, \dots, \mathbf{x}_{n-1}\}$. For each stack of feature vectors, an RNN with Gated Recurrent Units (GRU) is used to progressively generates prediction signals. This process is formulated as follows,
\begin{equation}
\begin{split}
\mathbf{z}_t =& \sigma(\mathbf{W^zx}_t + \mathbf{U^zh}_{t-1}), \\
\mathbf{r}_t =& \sigma(\mathbf{W^rx}_t + \mathbf{U^rh}_{t-1}), \\
\mathbf{h}_t =& \mathbf{z}_t\odot \mathbf{h}_{t-1} + (1-\mathbf{z}_t)\odot\\& \sigma(\mathbf{Wx}_t+\mathbf{U}(\mathbf{r}_t\odot \mathbf{h}_{t-1})+\mathbf{b}),
\end{split}
\end{equation}
where $\odot$ denotes element-wise multiplication of tensors and $\sigma$ is the non-linear activation function, which is \textit{tanh} in practice. At the step \textit{t}, the recurrent unit takes the \textit{t}-th feature vector $\mathbf{x}_t$ in the sequence as the input data. It also takes in the response in the last step, noted as $\mathbf{h}_{t-1}$. With learned parameters $\mathbf{W^z}$ for the input $\mathbf{x}_t$ and $\mathbf{U^z}$ for previous response $\mathbf{h}_{t-1}$, it produces the \textit{update gate} $\mathbf{z}_t$ to apply for the two source of data. Intuitively, when \textit{t} falls in the range of the valid reference samples, $\mathbf{x}_t$ contains useful information about the context. The gate weight indicates more confidence for $\mathbf{x}_t$ than $\mathbf{h}_{t-1}$ to better model the context information. When the unit is required to do prediction, \textit{i.e.} the input $\mathbf{x}_t$ contains much less information, the parameters $\mathbf{W^z}$ and $\mathbf{U^z}$ control the units to propagate the information encoded in $\mathbf{h}_{t-1}$. The \textit{reset gate} $\mathbf{r}_t$ is in the same form as the \textit{update gate}, but another pair of parameters are trained to control whether to reset the current state. Finally, the input data and the propagated data are fused to form the response of the \textit{t}-th step.
\RR{
The combined horizontal and vertical RNNs can handle complex textures. To use horizontal and vertical RNNs jointly, the textures in other directions can be modeled. The GRU (basic cell of our RNNs) includes reset and update gates, which enable to capture both short-term and long-term spatial dependency. Thus, it is good at modelling piecewise smooth regions and non-stationary textures. By stacking multiple spatial RNNs, PS-RNN can percept context information from a large region and is capable to predict textures along any direction with rather complex structures.}
After the progressive generation, the convolutional fusion component of the PS-RNN unit learns a robust merging of predicted feature maps. The feature vectors are concatenated back into feature maps and merged using a convolutional layer. As we split the progressive prediction into the horizontal and the vertical forms, we exploit the fusion component to help merge the results of the two predictors and achieve accurate prediction for complex textures.

\textbf{Analysis}.
We propose PS-RNN to address the drawbacks of previous CNNs and FCs, which directly predict missing pixels based on local features directly. First, with the update gate, the network can \textit{aggregate knowledge} selectively. It learns structural correlations in the reference area and generates predictions without the interference of the non-informative pixels in the to-be-predicted area. Second, with the reset gates, the network is able to generate non-linear complex textures like rings, as it is capable of \textit{forgetting previous knowledge} and restarting when transitioning to a new context, \textit{e.g.} new type of texture or new area.

To make this idea clear, we evaluate the ability for the proposed model to handle complex conditions with a visualization. As shown in Fig. \ref{fig:complex}, we visualize the prediction results of the network with the corresponding context. In Comparison I, we show that the proposed PS-RNN units can better utilize global contexts. HEVC (planar mode chosen by RDO) results in blurry predictions. In Comparison II and III, it is illustrated that noise in the texture can severely decrease the performance of directional prediction scheme in HEVC, while the proposed method can reduce the interference of the noise. In Comparison IV we can see that HEVC cannot handle textures consisting of two directional patterns, while PS-RNN networks are able to generate accurate predictions. 
}

\begin{figure*}[htbp]
\centering
\captionsetup{labelfont={color=black},font={color=black}}
    \begin{subfigure}[h]{0.02\linewidth}
    \centering
    \RRR{I}
    \end{subfigure}
    \begin{subfigure}[h]{0.18\linewidth}
    \centering
      \includegraphics[width=0.9\linewidth]{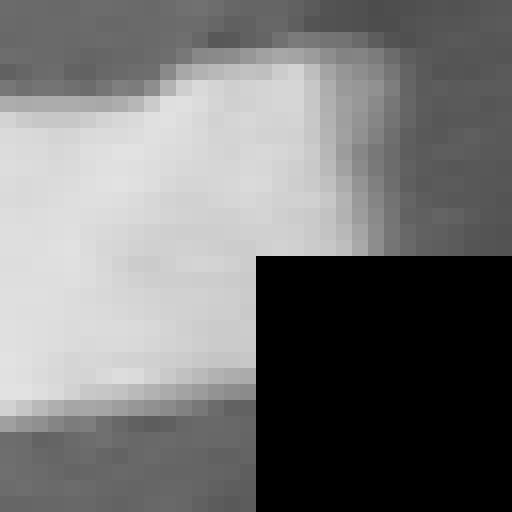}
    \end{subfigure}
    \begin{subfigure}[h]{0.18\linewidth}
    \centering
      \includegraphics[width=0.9\linewidth]{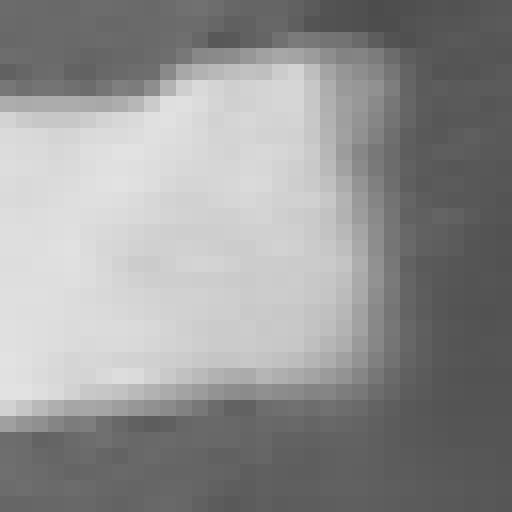}
    \end{subfigure}
    \begin{subfigure}[h]{0.18\linewidth}
    \centering
      \includegraphics[width=0.9\linewidth]{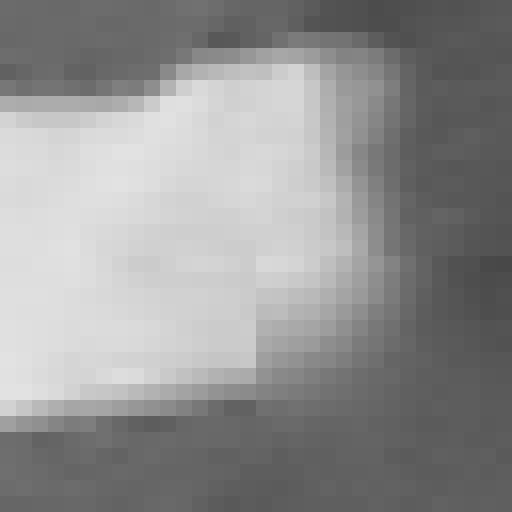}
    \end{subfigure}
    \begin{subfigure}[h]{0.18\linewidth}
    \centering
      \includegraphics[width=0.9\linewidth]{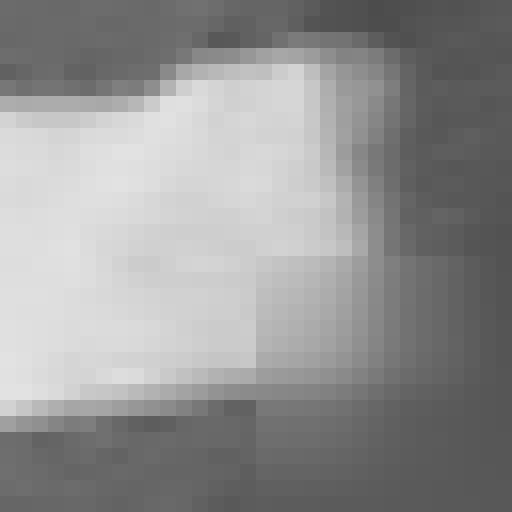}
    \end{subfigure}
    \begin{subfigure}[h]{0.18\linewidth}
    \centering
      \includegraphics[width=0.9\linewidth]{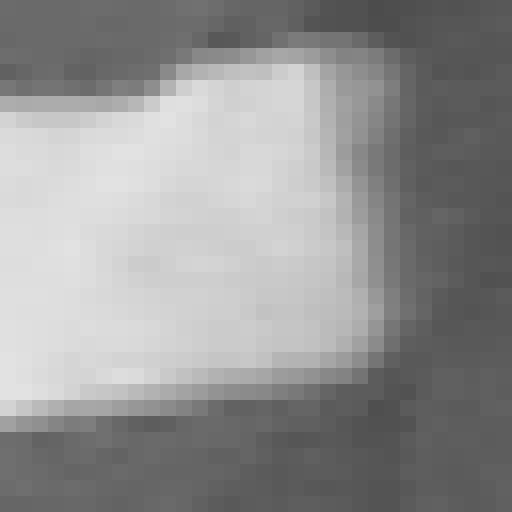}
    \end{subfigure}
    
    \vspace{10mm}
    
    \begin{subfigure}[h]{0.02\linewidth}
    \centering
    \RRR{II}
    \end{subfigure}
    \begin{subfigure}[h]{0.18\linewidth}
    \centering
      \includegraphics[width=0.9\linewidth]{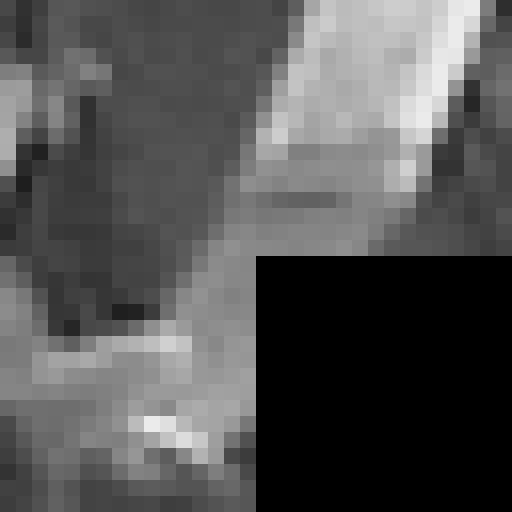}
    \end{subfigure}
    \begin{subfigure}[h]{0.18\linewidth}
    \centering
      \includegraphics[width=0.9\linewidth]{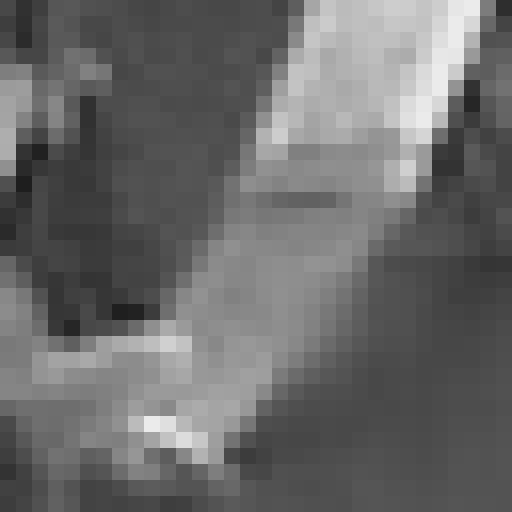}
    \end{subfigure}
    \begin{subfigure}[h]{0.18\linewidth}
    \centering
      \includegraphics[width=0.9\linewidth]{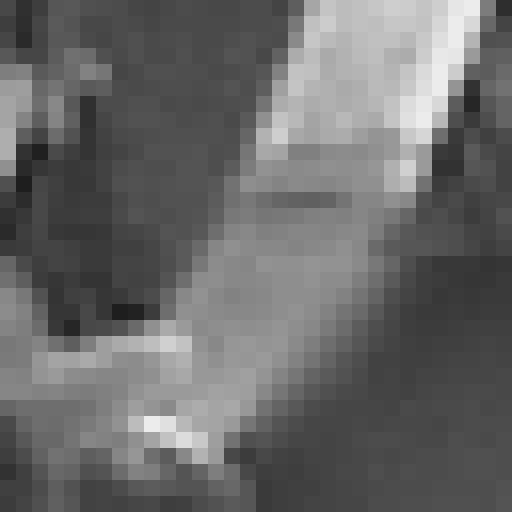}
    \end{subfigure}
    \begin{subfigure}[h]{0.18\linewidth}
    \centering
      \includegraphics[width=0.9\linewidth]{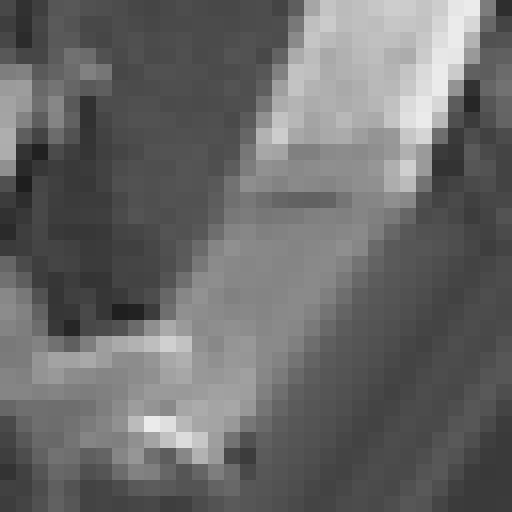}
    \end{subfigure}
    \begin{subfigure}[h]{0.18\linewidth}
    \centering
      \includegraphics[width=0.9\linewidth]{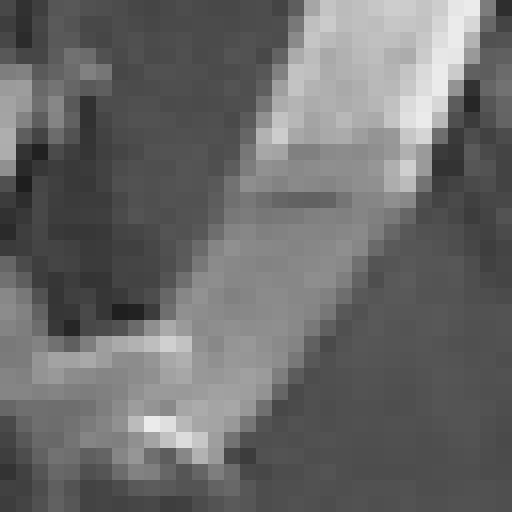}
    \end{subfigure}

    \vspace{10mm}
    
    \begin{subfigure}[h]{0.02\linewidth}
    \centering
    \RRR{III}
    \end{subfigure}
    \begin{subfigure}[h]{0.18\linewidth}
    \centering
      \includegraphics[width=0.9\linewidth]{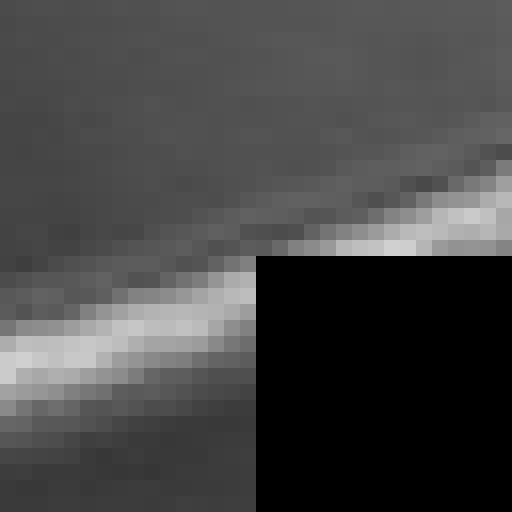}
    \end{subfigure}
    \begin{subfigure}[h]{0.18\linewidth}
    \centering
      \includegraphics[width=0.9\linewidth]{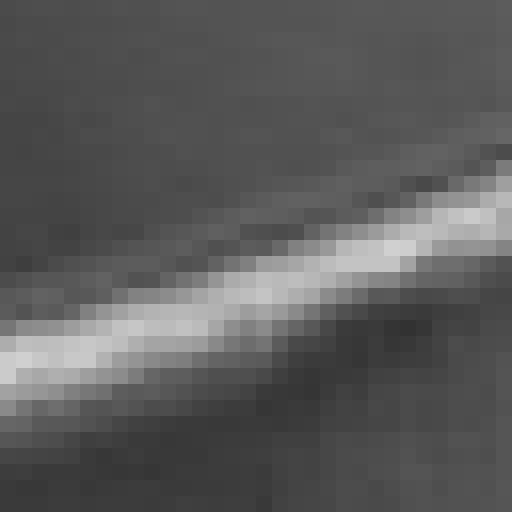}
    \end{subfigure}
    \begin{subfigure}[h]{0.18\linewidth}
    \centering
      \includegraphics[width=0.9\linewidth]{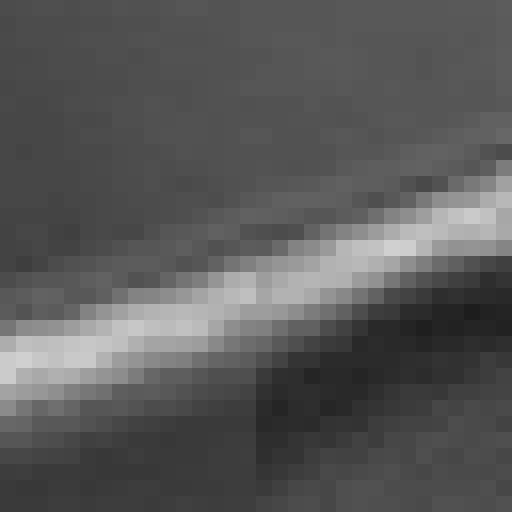}
    \end{subfigure}
    \begin{subfigure}[h]{0.18\linewidth}
    \centering
      \includegraphics[width=0.9\linewidth]{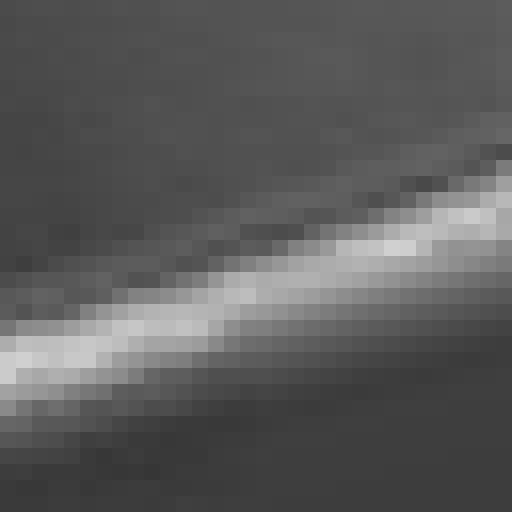}
    \end{subfigure}
    \begin{subfigure}[h]{0.18\linewidth}
    \centering
      \includegraphics[width=0.9\linewidth]{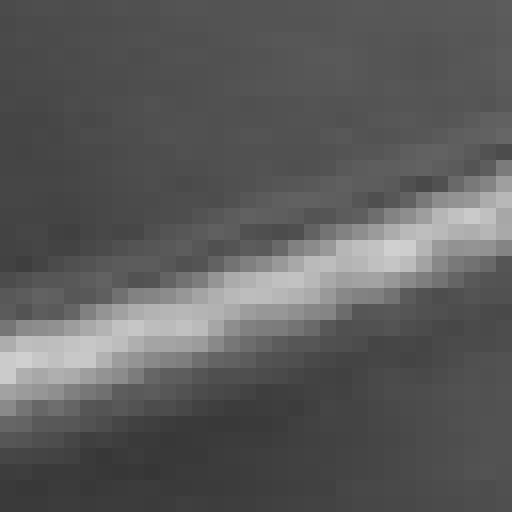}
    \end{subfigure}

    \vspace{10mm}
    
    \begin{subfigure}[h]{0.02\linewidth}
    \centering
    \RRR{IV}
    \end{subfigure}
    \begin{subfigure}[h]{0.18\linewidth}
    \centering
      \includegraphics[width=0.9\linewidth]{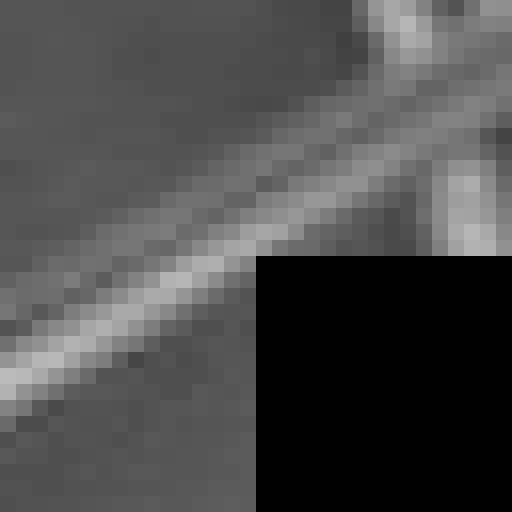}
      \caption{\RRR{Context}}
    \end{subfigure}
    \begin{subfigure}[h]{0.18\linewidth}
    \centering
      \includegraphics[width=0.9\linewidth]{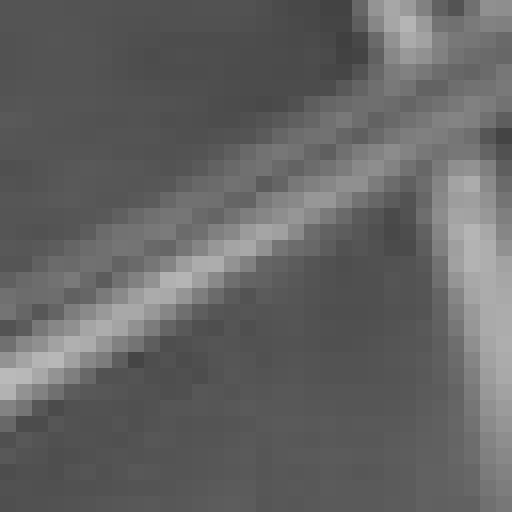}
      \caption{\RRR{PS-RNN}}
    \end{subfigure}
    \begin{subfigure}[h]{0.18\linewidth}
    \centering
      \includegraphics[width=0.9\linewidth]{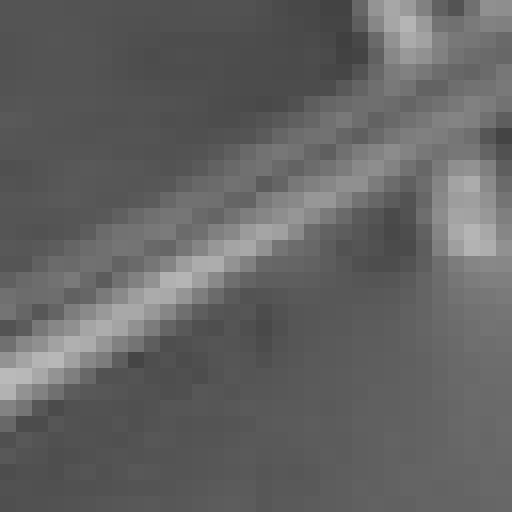}
      \caption{\RRR{FC}}
    \end{subfigure}
    \begin{subfigure}[h]{0.18\linewidth}
    \centering
      \includegraphics[width=0.9\linewidth]{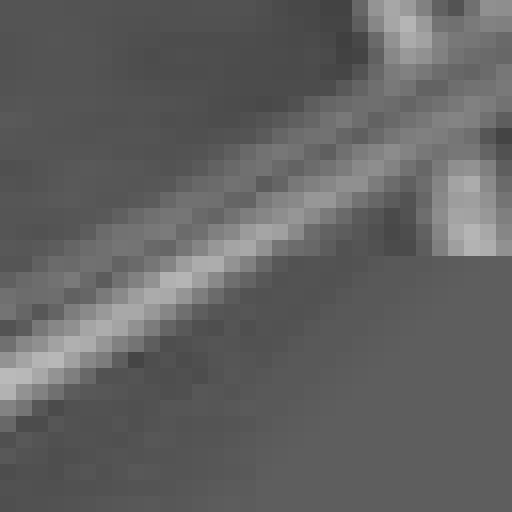}
      \caption{\RRR{HEVC}}
    \end{subfigure}
    \begin{subfigure}[h]{0.18\linewidth}
    \centering
      \includegraphics[width=0.9\linewidth]{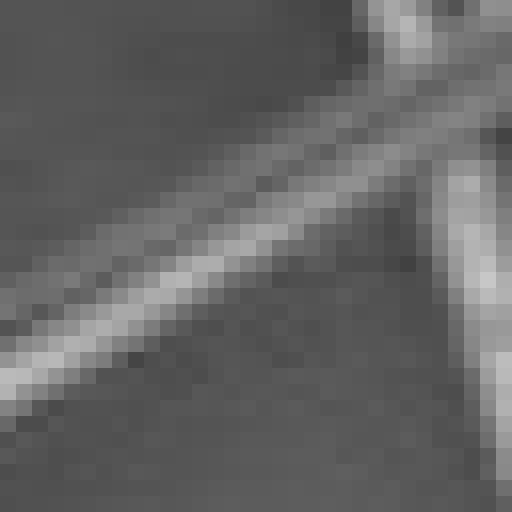}
      \caption{\RRR{Ground Truth}}
    \end{subfigure}

    \caption{\RRR{Visualization of the predictions produced by PS-RNN. PS-RNN produces the structural patterns accurately even without RDO process.}}
    \label{fig:complex} 
\end{figure*}

\RR{
\textbf{Implementation Detail}
We use GRU as the cell in the PS-RNN units. For the first PS-RNN unit, 8 cells are used for the vertical and the horizontal RNNs, respectively. For the other two units, we use 4 cells in the GRU. We train the network using Adam optimizer. PReLU is utilized as the activation function for the convolutional layers and we follow the original activation setting of GRU in the PS-RNN units. No activation function is used for the last layer and the result is clipped to $[0,1]$ for normalization. The initial learning rate is set to 0.001. We conduct a step-wise decay of the learning rate with the ratio 0.1 at the iterations of \{50,000, 75,000, 85,000\}, and we train the network for 100,000 iterations. We choose the model of the lowest validation loss of the last 20,000 iterations.}

\RR{
\subsection{SATD Loss Function}
\RRR{
We employ SATD as a loss function for training the proposed network. The core idea of SATD, namely Hadamard transform, is first proposed in \cite{pratt1969hadamard} for image compression. It redistributes the image energy properly and it can be efficiently implemented. Therefore, it is employed by many image and video codecs. In HEVC, SATD with Hadamard transform is used to evaluate rate-distortion costs. In the following, we first describe the key component of its calculation -- Hadamard transform, and then analyze SATD as a metric and loss function, respectively. Finally, we show the formulation of using SATD in back-propagation network training.}

\textbf{Hadamard Transform}. Different from MSE, in computing SATD, Hadamard transform is first applied to the difference of the predicted block and ground truth, namely the residue, before the difference is summed up as a loss. It is shown in \cite{kunz1979equivalence} that Hadamard transform is equivalent to discrete Fourier transform when applied to such data in intra prediction. Thus, we are able to separate the low-frequency component and the high-frequency component in the residue by evaluating using SATD. With such techniques, we can differently handle the low-frequency and the high-frequency component respectively. As this form of transformation is fast and easy to implement, it is used broadly in HEVC in the RDO process.

\textbf{SATD as a Metric}. In a transform coding scheme, the goal is to minimize the cost of encoding the residue between the original signal and the predicted one. In such a circumstance, the MSE as a metric has drawbacks in evaluating the actual cost of the encoding process. As the MSE sums up all the errors for each pixel regardless of its position, it is a spatially independent metric, while in the transform coding, the correlation of adjacent predicted pixels matters a lot. As a result, SATD is used to evaluate the rate-distortion cost for encoding the video. In computing the SATD of a residue block, the correlations of pixels are taken into consideration to reflect the cost to encode such a block with transform coding. Thus, SATD is more consistent with the final rate-distortion cost in the codec. 
 
\textbf{SATD as Loss Function}. As SATD is a better metric for the performance of our model to be used in the problem of intra prediction, we propose to use SATD loss function to train the network for the task. To adopt SATD for network training, we have a closer examination of the calculation of SATD.} We define $\mathbf{D} = \tilde{\mathbf{Y}}-\mathbf{Y}$ as the difference between the prediction $\tilde{\mathbf{Y}}$ and the ground truth $\mathbf{Y}$. Without loss of generality, only the condition where the shape of the Hadamard transform matrix $\mathbf{H}$ is the same as $\mathbf{D}$ is illustrated. \YY{When $\mathbf{D}$ is} larger in shape, it can be partitioned before the transformation. We apply the transform as,
\begin{equation}
\mathbf{D}' = \mathbf{HDH}^T,
\end{equation}
where $\mathbf{D}'$ is the transformed difference. Note that the matrix $\mathbf{H}$ is symmetric, $\mathbf{D}'$ can also be expressed as,
\begin{equation}
\mathbf{D}' = \mathbf{H}\mathbf{D}\mathbf{H}.
\end{equation}
\YY{We define the SATD} loss function $S$ as,
\begin{equation}
\label{equ_SATD1}
S = \|\mathbf{HDH}\|_1=\sum_i\sum_j|\mathbf{D}'_{ij}|.
\end{equation}
The main drawback of MSE in the back-propagation process is that the derivative of $S$ in MSE loss function with respect to $\mathbf{D}_{ij}$ is an expression only related to $\mathbf{D}_{ij}$ itself. \RRR{As a consequence, the correlations between this pixel and its neighboring pixels are not measured. In the proposed method, rather than using MSE as the metric, we use SATD to measure such correlations to control the bit-rate for encoding the transformed difference $\mathbf{D'}$.} Thus, the network is trained towards the goal of optimizing the rate-distortion performance of the codec.

As the network is trained using the back-propagation method, we calculate the partial derivative of the loss $S$ with respect to each entry of the distance $\mathbf{D}$ as. The detailed derivation for the back-propagation is presented in the Appendix. We further illustrate the effect for SATD loss function to better benefits the network to improve the rate-distortion performance in Section \ref{experimental-results}.

\subsection{Variable Block Size}

In HEVC, using larger block size saves the bit-rate, as fewer bits are needed to encode block-level flags. Supporting more flexible coding structure is one source of bit-rate reduction of HEVC compared with previous codecs. In directional intra prediction, a large block with simple texture is directly predicted to save bit-rate. However, previous work \cite{li2017icip} on employing deep learning in image and video compression restricts the block size for both the anchor and the proposed model to relatively small scale like $8\times8$. With this restriction, the performance gain of the proposed method in real coding conditions is not fully unveiled.

In our work, variable block size is supported. We allow the size of PU to be from $4\times4$ to $32\times32$, controlled by the split strategy of HEVC. To train the network for diverse contents scale, we train the network on a diversified dataset where frames of different scale are mixed. \RR{In HEVC, all block-sizes share the same intra prediction scheme. With the limitation of the directional prediction scheme, HEVC tends to split a large block to smaller ones when its textures are complicated, where more bits are used to encode the prediction mode due to the split. Our proposed model can generate more accurate predictions for large blocks compared with HEVC.}

\subsection{Integration with the HEVC}

We implement the network in HEVC Test Model (HM) 16.15. Due to the diversity of the video content, it is hard for a predictor to handle all the conditions. \YY{Instead of totally} replacing the intra prediction part or overwriting one specific directional mode for the intra prediction of HEVC, we use RDO to decide whether to use the original HEVC predictor or the proposed predictor. For cases where the original predictor can perform well enough, RDO chooses the original HEVC scheme. For other cases where HEVC fails, the network is selected to handle complex texture. One additional flag is added to each PU to indicate the selection. \YY{No flag for the directional mode} is needed when the proposed model is selected.

\begin{figure}[htbp]
    \centering
    \includegraphics[width=1.0\linewidth]{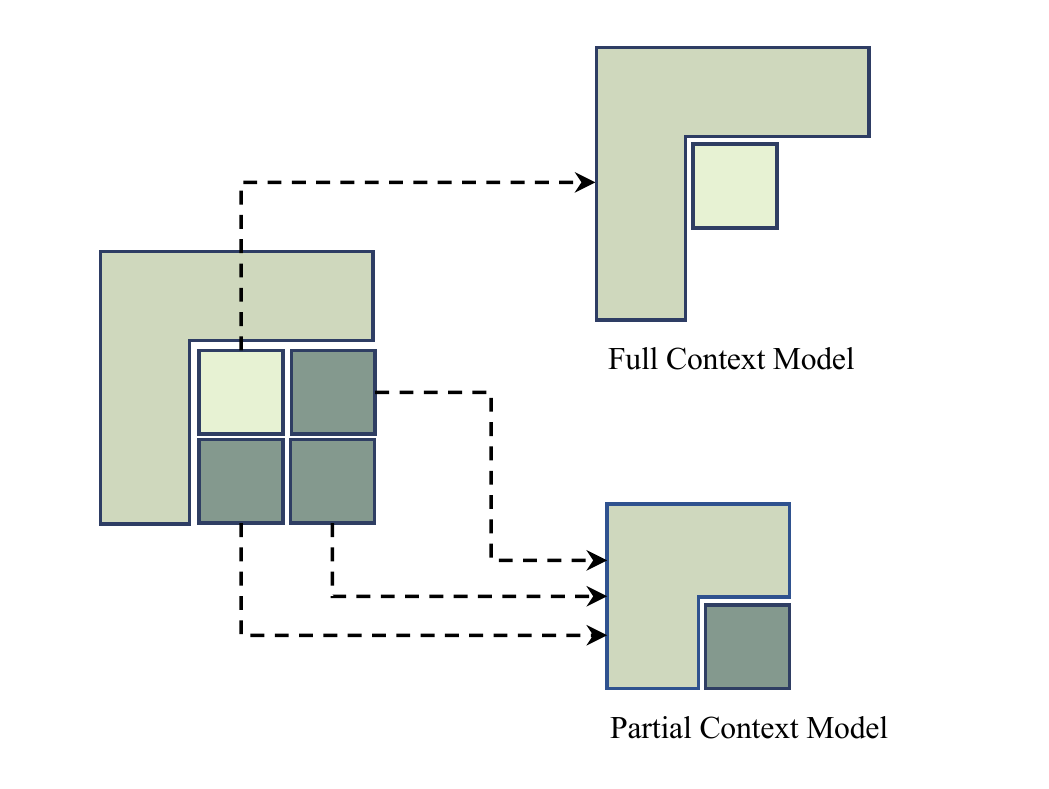}
    \caption{Different availability of reference samples in a coding unit. Blocks with two different colors are processed using two different models.}
    \label{fig:refmode}
\end{figure}

Though the reconstructed pixels on the left below of the current PU can be used as reference samples, they are not always available. Fig. \ref{fig:refmode} shows an example. Only \YY{one of four split PUs} can have full reference samples as we mentioned above. \YY{Other three} PUs only have limited reference blocks. To address this issue, \YY{the missing pixels for reference are simply filled with DC value in HEVC.} In our deep model based intra prediction, filling blocks with constant value brings in extra complexity. Redundant information is fed in the network, which results in poor performance. Thus, we separately train two models for these two conditions. The top-left PU is predicted by a model which is trained using full context. The rest is predicted by a model for the three-block condition where the blocks from above, the left and left-above are used as training context.

\section{Experimental Results}\label{experimental-results}

\subsection{Training Settings}

In data-driven methods, training materials are of great importance. As video frames are quite similar in one sequence, they are not ideal training data for intra prediction models, where content diversity can benefit model training. To effectively train the network, in our proposed method, the training data is generated from high-resolution images provided in \cite{Timofte_2017_CVPR_Workshops}. These images include a wide range of contents, including natural view and artificial scenery. They are diversified in texture, color, and brightness which benefits the network training.

We propose the model to work for various resolutions, so we train the model with materials of various scales. The images are cropped and downsampled to three scales, namely $1792\times1024$, $1344\times768$, and $896\times512$. Using these images with different scales, our network can work for videos from high resolution to low resolution. Further, to reduce the gap between the distribution of the training set and the test set, the images are previously encoded using HEVC. We set the Quantization Parameter (QP) to $22, 27, 32, 37$ and use the reconstructed blocks in the decoding process to form the training pairs. We randomly sample about 3,000,000 pairs to train the model. A training process takes about 4 hours on an NVIDIA GTX 1080 GPU. Adam optimizer \cite{kingma2014adam} is used for training. The network is implemented using TensorFlow \cite{tensorflow2015-whitepaper}.

Training the model using pairs with high QP settings can enhance the ability of models to mitigate the influence of the quantization noise, but these training pairs are less expressive. To enhance the robustness to noise while avoiding over-smoothing, the training material is mixed with reconstructed samples in low and high QPs. 

\begin{figure}[htbp]

\captionsetup{labelfont={color=black},font={color=black}}
    \centering
    \begin{subfigure}[b]{1.0\linewidth}
        \centering
    \includegraphics[width=1\linewidth]{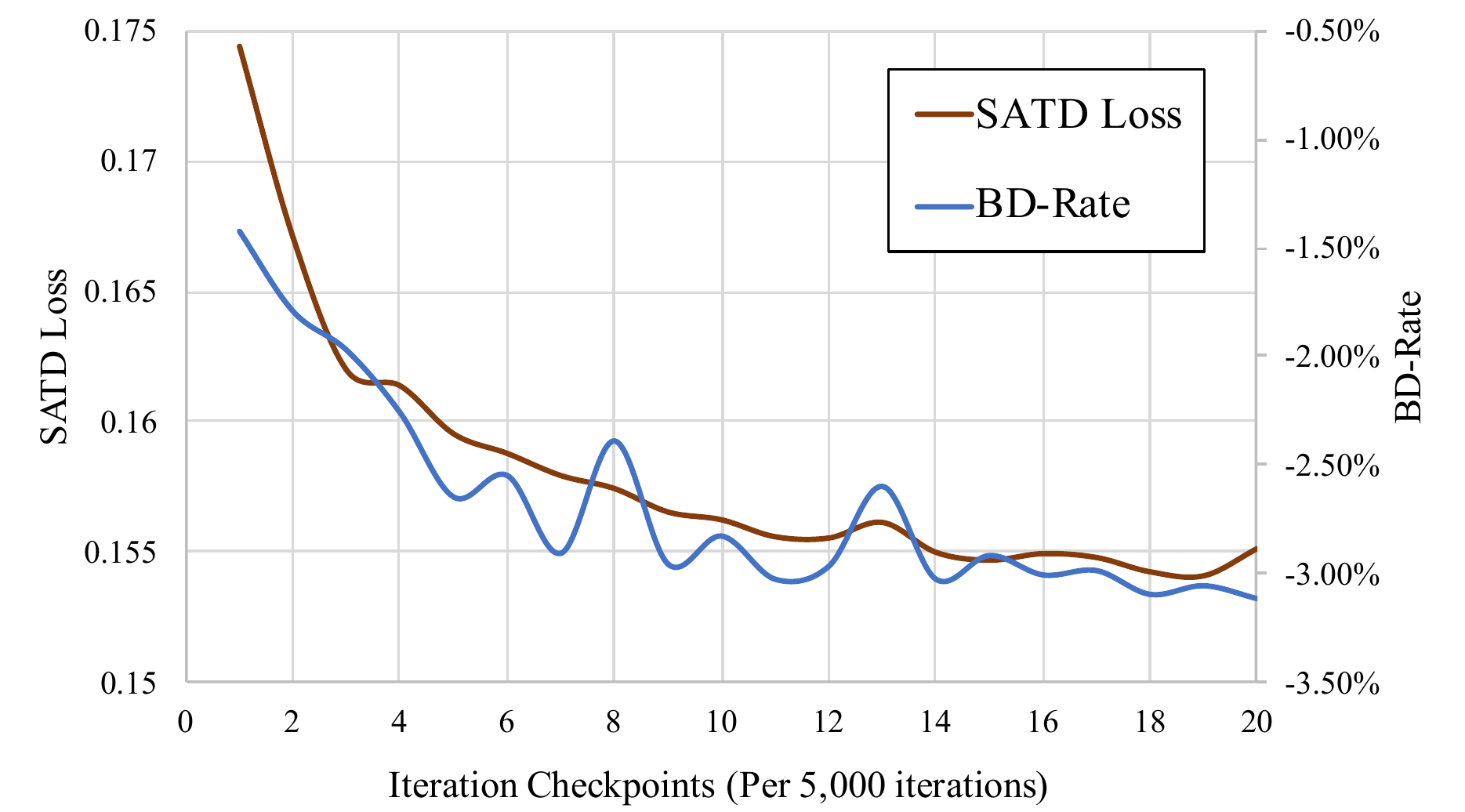}
    \caption{\RR{Values of SATD loss together with the corresponding BD-Rate on the validation set during the training process of the network.}}
    \label{fig:satd_rd}
  	\end{subfigure}
  	\begin{subfigure}[b]{1.0\linewidth}
        \centering
    \includegraphics[width=1\linewidth]{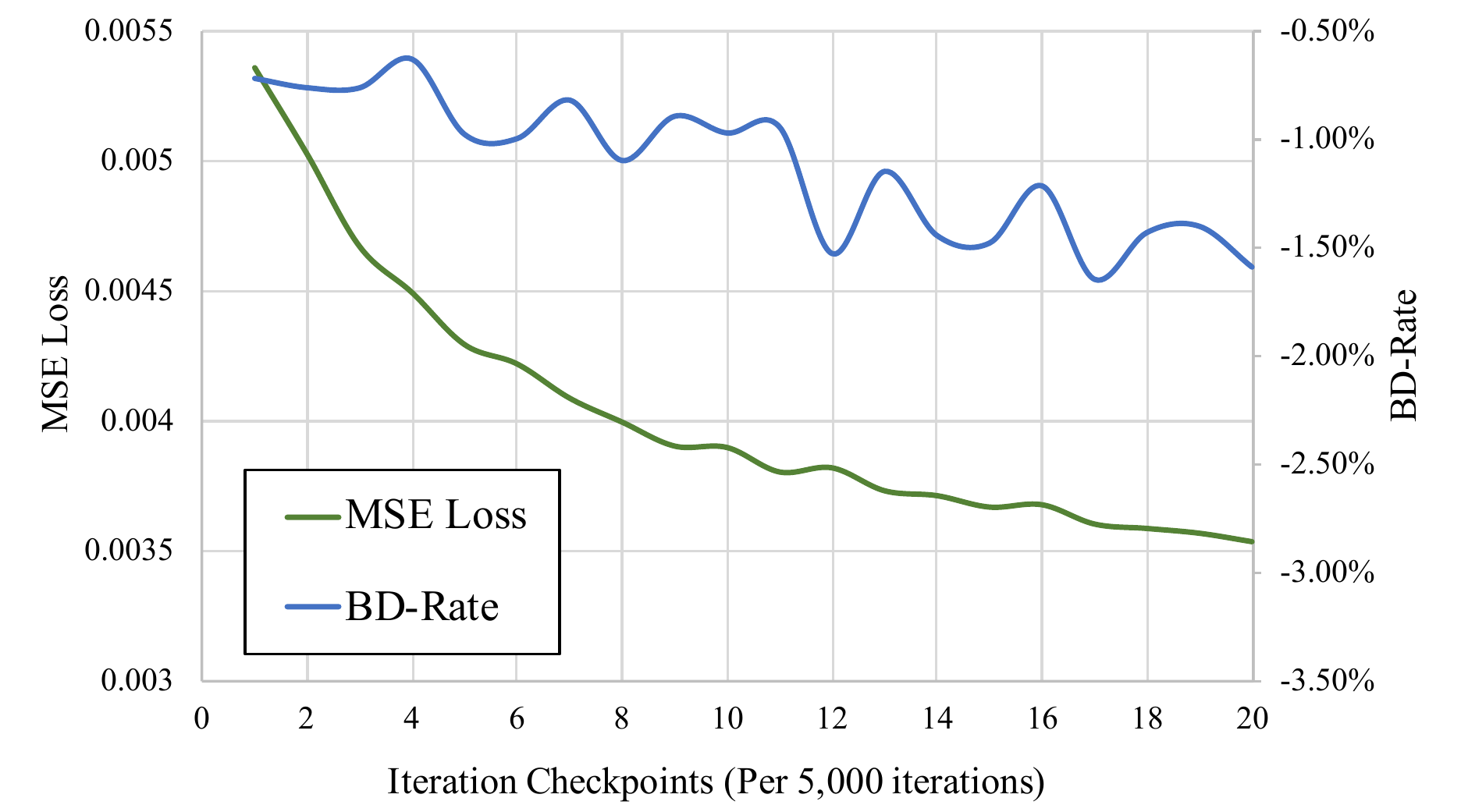}
    \caption{\RR{Values of MSE loss together with the corresponding BD-Rate on the validation set during the training process of the network.}}
    \label{fig:mse_rd}
  	\end{subfigure}
    \caption{\RR{Comparison of SATD and MSE loss function corresponding to the rate-distortion performance in HEVC. First, networks trained with SATD converge faster and perform better than those trained with MSE. Second, the variation of the evaluated BD-Rate is larger in networks trained using MSE loss function, which implies the incapability of MSE loss to indicate the actual rate-distortion performance of the trained models in the codec.}}

	\label{fig:satd_mse_rd} 
\end{figure}

\RR{
\subsection{Effectiveness of SATD Loss}

The goal of intra prediction is to improve the rate-distortion performance of video coding. SATD is able to measure the bitrate and the distortion jointly, thus it is more commonly used for video coding. We compare the models trained using SATD and MSE respectively. We first illustrate the variation of the loss and the corresponding BD-Rate evaluated on the validation set during the training process in Fig. \ref{fig:satd_mse_rd}. The statistics of the loss value and the corresponding rate-distortion performance indicate the superiority of SATD to MSE as a loss function for intra prediction. First, networks trained with SATD converge faster and perform better than those trained with MSE. The final performance of SATD-trained models is far better than that of MSE-trained models. Second, the variation of the evaluated BD-Rate is larger in networks trained using MSE loss function, which implies the incapability of MSE loss to indicate the actual rate-distortion performance of the trained models in the codec. We also evaluate the overall performance of PS-RNN models trained using MSE and SATD as the loss function respectively. As shown in TABLE \ref{tab:satd}, with the same architecture, networks trained with SATD have superior performance in BD-Rate to that of networks trained with MSE. We also apply SATD loss function to FC networks \cite{li2017icip}. We adopt the same setting used in \cite{li2017icip} and we can see that SATD loss function brings in significant improvement on FC networks. The experimental results show that SATD is an effective loss function for different networks in intra prediction, and it leads to a better performance than MSE in intra prediction.
}

\subsection{Evaluation of Recurrent Structure}

To evaluate the effectiveness of the progressive recurrent structure of the proposed network, we also compare our PS-RNN model with FC networks. To conduct this comparative experiment, we implement an eight-layer FC network (FC-SATD) which has approximately the same amount of parameters as our proposed PS-RNN. Note that in our experiment, to make the comparison fair, both the FC model and PS-RNN model are trained using SATD loss function. The comparison results are shown in BD-Rate \cite{bjontegarrd2001calculation} in Table \ref{tab:satd}. We also compare the results with the FC network in \cite{li2017icip}, where the same $8\times8$ block size restriction is applied and MSE is used as the loss function. As we can see from the quantitative result, by introducing the SATD loss function, we achieve a leap in performance for intra prediction using the same FC architecture. We further improve the performance with our proposed PS-RNN network.

\begin{figure}[htbp]

\captionsetup{labelfont={color=black},font={color=black}}
    \centering
    \includegraphics[width=1.0\linewidth]{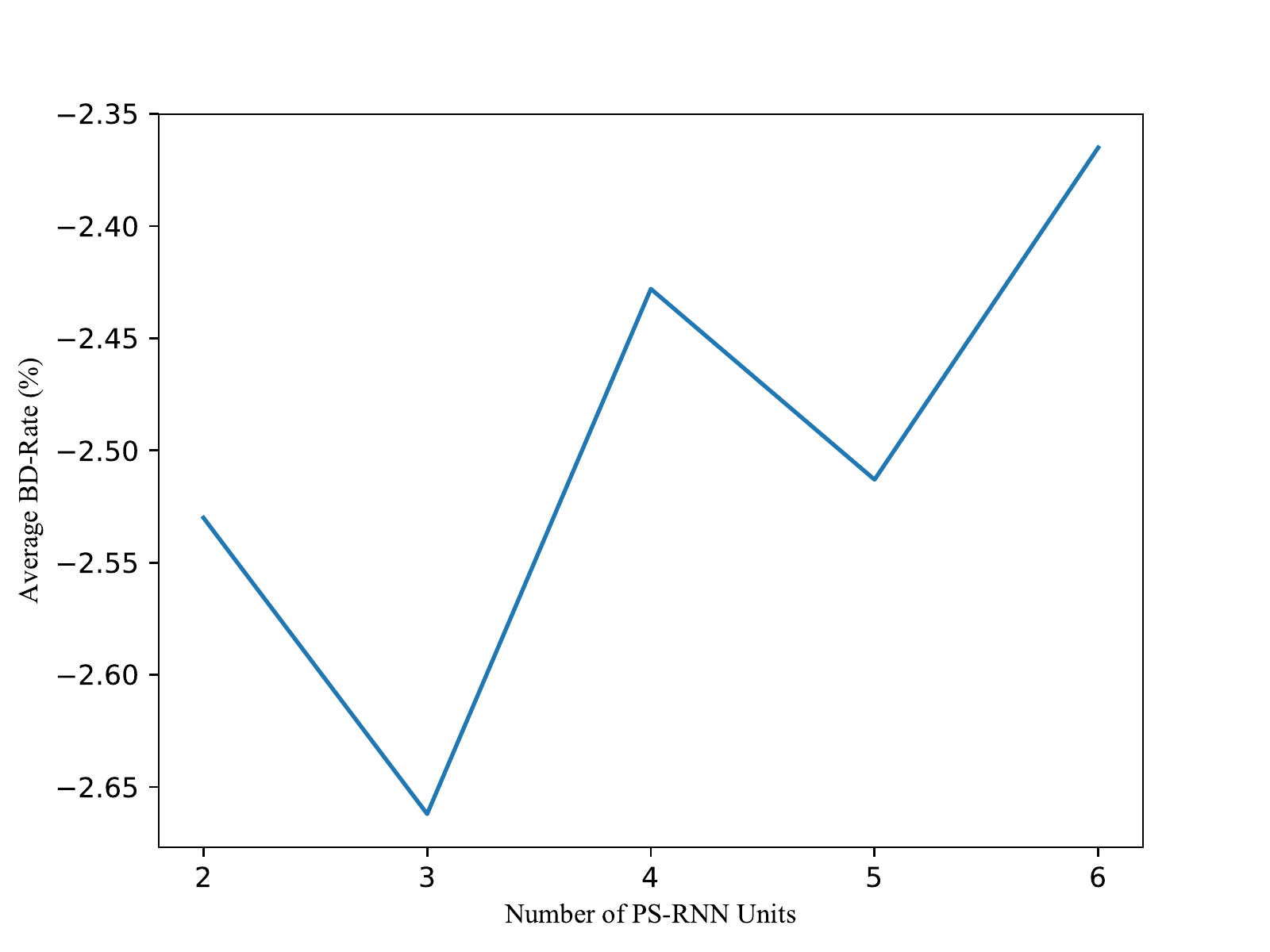}
    \caption{\RRR{The evaluation of average BD-Rate with a different number of PS-RNN units.}}
    \label{fig:unit_rd}
\end{figure}

\begin{table*}[htbp]
\centering
\caption{Quantitative analysis of selected methods. The results are shown in BD-Rate using HEVC (HM 16.15) as the anchor. PU size is set to $8\times8$ in both the proposed model and the anchor.}
\label{tab:satd}
\scalebox{1.1}{
\begin{tabular}{c|c|c|c|c|c}
\hline
Class & Sequence & PS-RNN-SATD & PS-RNN-MSE & FC-SATD & Li \cite{li2017icip} \bigstrut\\
\hline
\multirow{5}[4]{*}{Class A} & Traffic & \textbf{-3.8\%} & -2.3\% & -3.1\% & -1.0\% \bigstrut[t]\\
      & PeopleOnStreet & \textbf{-3.8\%} & -2.2\% & -3.1\% & -1.3\% \\
      & Nebuta(10bit) & \textbf{-1.9\%} & \textbf{-1.9}\% & \textbf{-1.9\%} & -1.6\% \\
      & SteamLocomotive(10bit) & \textbf{-3.2\%} & -2.8\% & \textbf{-3.2\%} & -1.7\% \bigstrut[b]\\
\cline{2-6}      & Class A Average & \textbf{-3.2\%} & -2.3\% & -2.8\% & -1.4\% \bigstrut\\
\hline
\multicolumn{1}{c|}{\multirow{6}[4]{*}{Class B}} & Kimono & \textbf{-6.6\%} & -3.6\% & -6.4\% & -3.2\% \bigstrut[t]\\
      & ParkScene & \textbf{-3.4\%} & -1.9\% & -2.9\% & -1.1\% \\
      & Cactus & \textbf{-3.3\%} & -1.8\% & -2.2\% & -0.9\% \\
      & BasketballDrive & \textbf{-7.8\%} & -3.2\% & -3.7\% & -0.9\% \\
      & BQTerrace & \textbf{-2.6\%} & -1.8\% & -1.6\% & -0.5\% \bigstrut[b]\\
\cline{2-6}      & Class B Average & \textbf{-4.7\%} & -2.5\% & -3.4\% & -1.3\% \bigstrut\\
\hline
\multicolumn{1}{c|}{\multirow{5}[4]{*}{Class C}} & BasketballDrill & \textbf{-2.9\%} & -1.5\% & -1.9\% & -0.3\% \bigstrut[t]\\
      & BQMall & \textbf{-2.9\%} & -1.9\% & -1.4\% & -0.3\% \\
      & PartyScene & \textbf{-2.3\%} & -1.8\% & -1.1\% & -0.4\% \\
      & RaceHorses & \textbf{-2.8\%} & -2.1\% & -2.3\% & -0.8\% \bigstrut[b]\\
\cline{2-6}      & Class C Average & \textbf{-2.7\%} & -1.8\% & -1.7\% & -0.5\% \bigstrut\\
\hline
\multicolumn{1}{c|}{\multirow{5}[4]{*}{Class D}} & BasketballPass & \textbf{-2.5\%} & -1.7\% & -1.4\% & -0.4\% \bigstrut[t]\\
      & BQSquare & \textbf{-1.8\%} & -1.2\% & -0.8\% & -0.2\% \\
      & BlowingBubbles & \textbf{-2.3\%} & -1.6\% & -1.7\% & -0.6\% \\
      & RaceHorses & \textbf{-2.6\%} & -2.5\% & -2.2\% & -0.6\% \bigstrut[b]\\
\cline{2-6}      & Class D Average & \textbf{-2.3\%} & -1.8\% & -1.5\% & -0.5\% \bigstrut\\
\hline
\multicolumn{1}{c|}{\multirow{4}[2]{*}{Class E}} & Johnney & \textbf{-6.8\%} & -3.8\% & -4.7\% & -1.0\% \bigstrut[t]\\
      & FourPeople & \textbf{-5.6\%} & -2.8\% & -4.1\% & -0.8\% \\
      & KristenAndSara & \textbf{-6.6\%} & -2.9\% & -4.0\% & -0.8\% \\
      & Class E Average & \textbf{-6.3\%} & -3.2\% & -4.3\% & -0.9\% \bigstrut[b]\\
\hline
\multicolumn{2}{c|}{Average} & \textbf{-3.8\%} & -2.3\% & -2.7\% & -0.9\% \bigstrut\\
\hline
\end{tabular}%
}
\end{table*}
\RR{
To allow non-linear generation of the prediction signals, we exploit multiple PS-RNN units to conduct progressive prediction for several times. To decide the optimal times of progressive prediction, we evaluate rate-distortion performance with a different number of the PS-RNN units. The results are shown in Fig. \ref{fig:unit_rd}. It is observed that, the network with three PS-RNN units achieves the best performance in BD-Rate. Thus, we choose to use three stacked PS-RNN units in the proposed method.
}

\subsection{Visualization Analysis}
\begin{figure}[htbp]

\captionsetup{labelfont={color=black},font={color=black}}
\centering
    \begin{subfigure}[h]{0.02\linewidth}
    \centering
    \RR{I}
    \end{subfigure}
    \begin{subfigure}[h]{0.31\linewidth}
    \centering
      \includegraphics[width=0.9\linewidth]{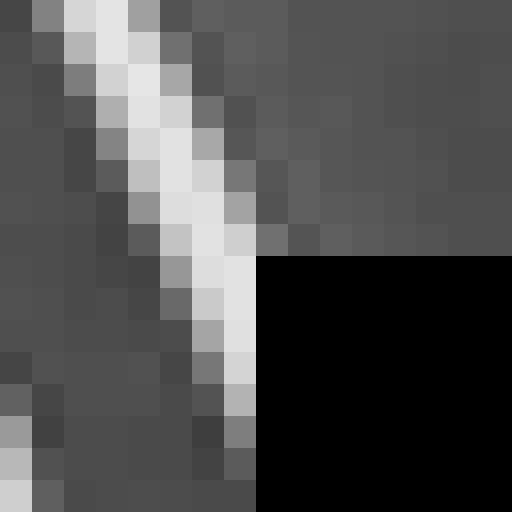}
    \end{subfigure}
    \begin{subfigure}[h]{0.31\linewidth}
    \centering
      \includegraphics[width=0.9\linewidth]{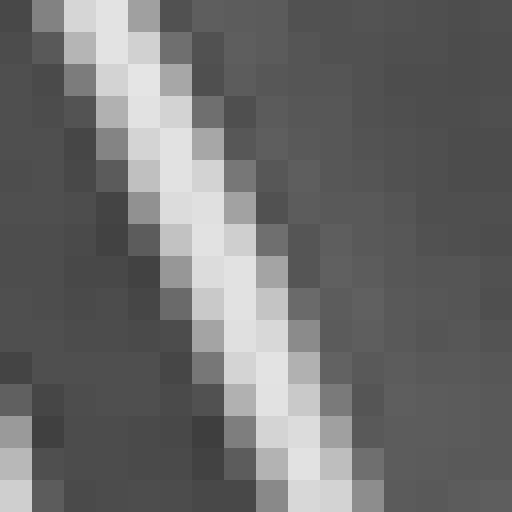}
    \end{subfigure}
    \begin{subfigure}[h]{0.31\linewidth}
    \centering
      \includegraphics[width=0.9\linewidth]{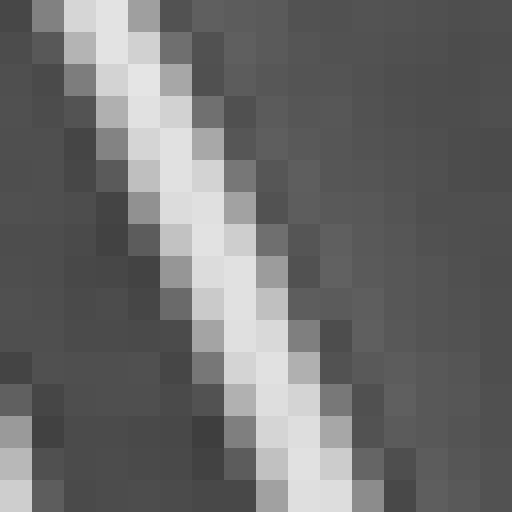}
    \end{subfigure}
    
    \vspace{4mm}
    
    \begin{subfigure}[h]{0.02\linewidth}
    \centering
    \RR{II}
    \end{subfigure}
    \begin{subfigure}[h]{0.31\linewidth}
    \centering
      \includegraphics[width=0.9\linewidth]{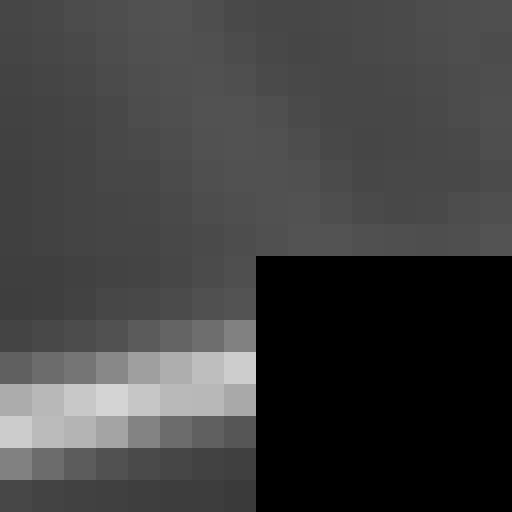}
    \end{subfigure}
    \begin{subfigure}[h]{0.31\linewidth}
    \centering
      \includegraphics[width=0.9\linewidth]{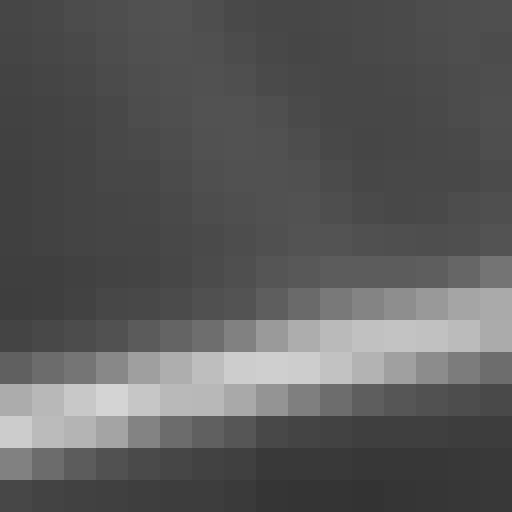}
    \end{subfigure}
    \begin{subfigure}[h]{0.31\linewidth}
    \centering
      \includegraphics[width=0.9\linewidth]{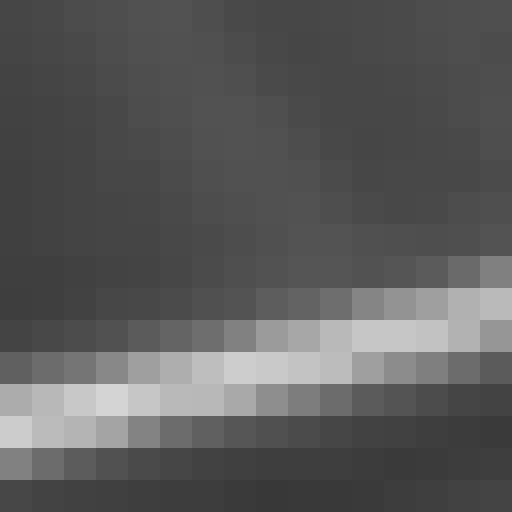}
    \end{subfigure}
    
    \vspace{4mm}
    
    \begin{subfigure}[h]{0.02\linewidth}
    \centering
    \RR{III}
    \end{subfigure}
    \begin{subfigure}[h]{0.31\linewidth}
    \centering
      \includegraphics[width=0.9\linewidth]{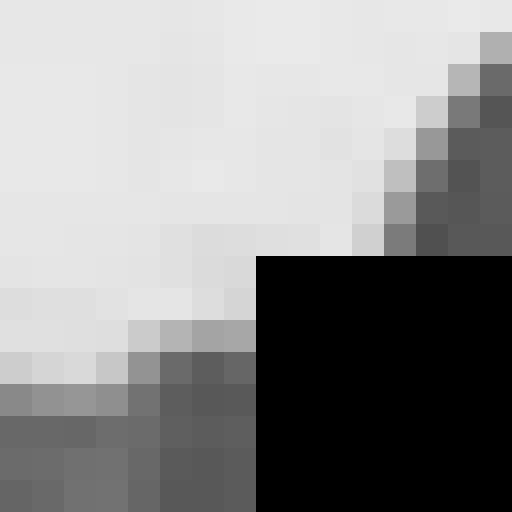}
    \end{subfigure}
    \begin{subfigure}[h]{0.31\linewidth}
    \centering
      \includegraphics[width=0.9\linewidth]{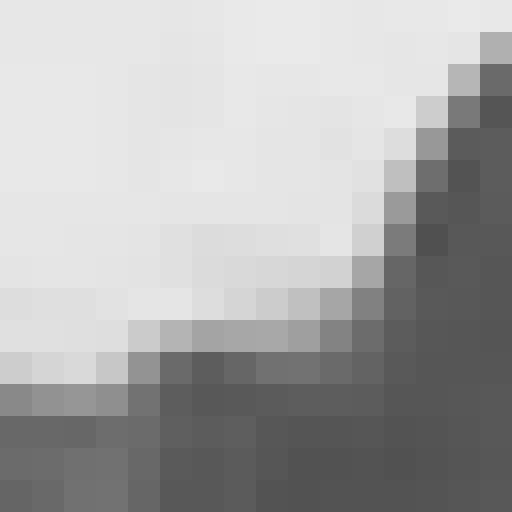}
    \end{subfigure}
    \begin{subfigure}[h]{0.31\linewidth}
    \centering
      \includegraphics[width=0.9\linewidth]{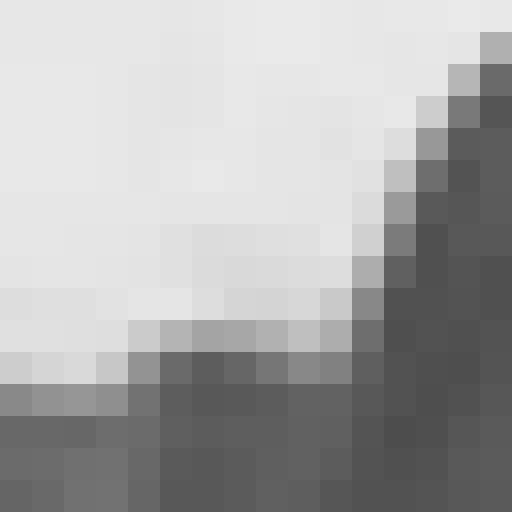}
    \end{subfigure}
    
    \vspace{4mm}
    
    \begin{subfigure}[h]{0.02\linewidth}
    \centering
    \RR{IV}
    \end{subfigure}
    \begin{subfigure}[h]{0.31\linewidth}
    \centering
      \includegraphics[width=0.9\linewidth]{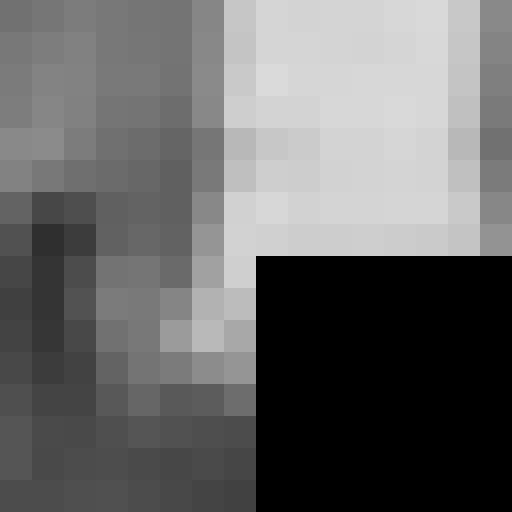}
      \caption{Context }
    \end{subfigure}
    \begin{subfigure}[h]{0.31\linewidth}
    \centering
      \includegraphics[width=0.9\linewidth]{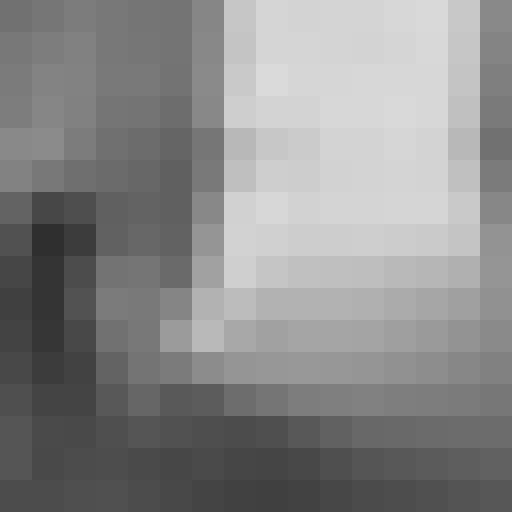}
      \caption{PS-RNN }
    \end{subfigure}
    \begin{subfigure}[h]{0.31\linewidth}
    \centering
      \includegraphics[width=0.9\linewidth]{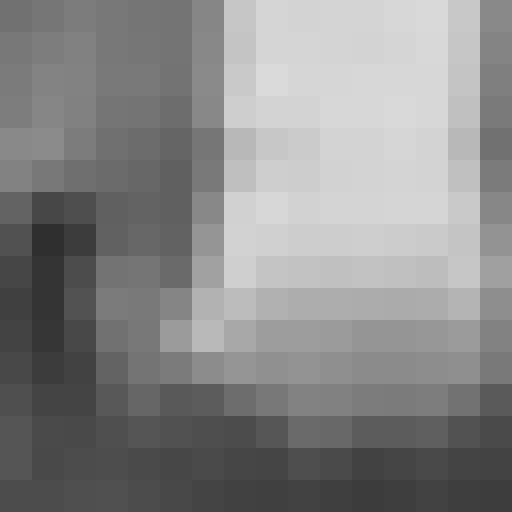}
      \caption{Ground Truth}
    \end{subfigure}
    
    \caption{\RR{Prediction results on natural context. As show in (I) and (II), our proposed model can handle directional texture without RDO. Also as illustrated in (III) and (IV), the proposed model can also tackle complex textures.}}
    \label{fig:vis_multi_nat}
\end{figure}

\RRR{
To investigate the source of improvement, we visualize the prediction signals of PS-RNN and FC network, as well as original HEVC intra prediction scheme in Fig. \ref{fig:complex}. It is shown in the figure that the proposed PS-RNN model can handle intra prediction in both directional and more complex conditions, without RDO of modes. The directional intra prediction scheme in HEVC can only handle directional pattern and flags for intra modes need to be transmitted. We also evaluate the predictions in more cases of natural images compared with FC models, as shown in Fig. \ref{fig:vis_multi_nat}. It is illustrated that the progressiveness of PS-RNN keeps the structure of the edges in the original image better than FC networks.
}

\begin{figure}[htbp]

\captionsetup{labelfont={color=black},font={color=black}}
\centering
    \begin{subfigure}[h]{0.48\linewidth}
    \centering
      \includegraphics[width=0.9\linewidth]{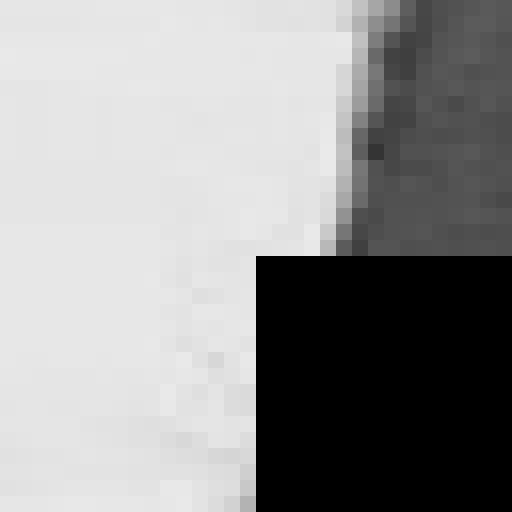}
      \caption{Context}
    \end{subfigure}
    \begin{subfigure}[h]{0.48\linewidth}
    \centering
      \includegraphics[width=0.9\linewidth]{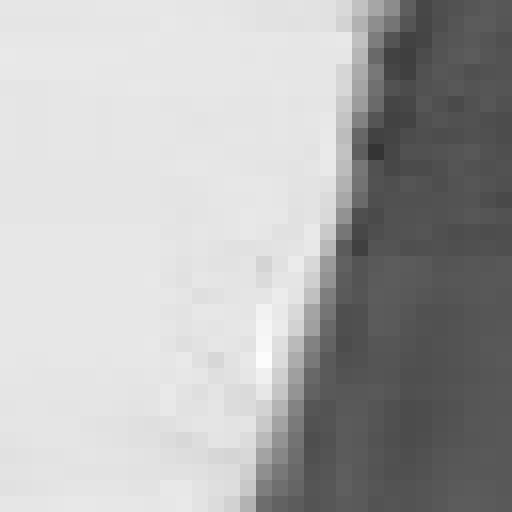}
      \caption{PS-RNN}
    \end{subfigure}
    \begin{subfigure}[h]{0.48\linewidth}
    \centering
      \includegraphics[width=0.9\linewidth]{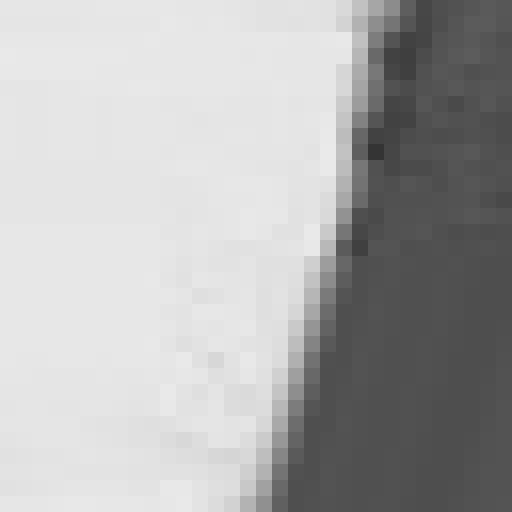}
      \caption{HEVC}
    \end{subfigure}
    \begin{subfigure}[h]{0.48\linewidth}
    \centering
      \includegraphics[width=0.9\linewidth]{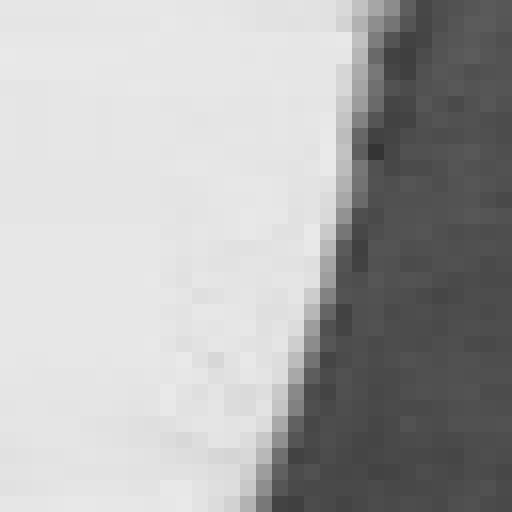}
      \caption{Ground Truth}
    \end{subfigure}
    \caption{\RR{Failure case analysis. (a) Intra prediction context. (b) Prediction results of PS-RNN network ($16\times16$). (c) Prediction results of HEVC (after RDO). (d) Original signal of the to-be-predicted block. HEVC can better handle sharp linear edges. \RRR{As PS-RNN is trained as a general model for all direction modes and the prediction without the RDO support is highly ill-posed, it tends to produce a \textit{safe} but blurry result.}}}
    \label{fig:vis_multi} 
\end{figure}

However, there are still artifacts in the prediction signal. A failure case analysis is presented in Fig. \ref{fig:vis_multi} The reason for this phenomenon, as we analyze, is that intra prediction is an ill-posed problem, where PS-RNN tends to generate the results which at a large probability lead to small residues guided by SATD loss, and may deviate from the ground truth slightly for the input samples.

\subsection{Variable Block Size Analysis}

\begin{table*}[htbp]
\captionsetup{labelfont={color=black},font={color=black}}
  \centering
  \caption{\RRR{Performance evaluation of PS-RNN and the unified integration PS-RNN+. The performance is evaluated with BD-Rate using the all-intra main configuration. The anchor is HM 16.15. The first frame of each sequence is tested.}}
  
\label{table:detail}
 \RRR{
 \scalebox{1.1}{
\begin{tabular}{c|c|c|c|c|c|c|c}

\hline
\multirow{2}[4]{*}{Class} & \multirow{2}[4]{*}{Sequence} & \multicolumn{3}{c|}{PS-RNN +} & \multicolumn{3}{c}{PS-RNN} \bigstrut\\
\cline{3-8}      &       & Y     & U     & V     & \multicolumn{1}{c|}{Y} & \multicolumn{1}{c|}{U} & \multicolumn{1}{c}{V} \bigstrut\\
\hline
\multicolumn{1}{c|}{\multirow{6}[12]{*}{Class B}} & Kimono & -2.21\% & -1.41\% & -1.05\% & -1.23\% & -0.87\% & -0.94\% \bigstrut\\
\cline{2-8}      & ParkScene & -2.83\% & -1.54\% & -1.29\% & -2.67\% & -1.58\% & -1.32\% \bigstrut\\
\cline{2-8}      & Cactus & -2.52\% & -1.28\% & -1.44\% & -2.34\% & -1.50\% & -0.88\% \bigstrut\\
\cline{2-8}      & BasketballDrive & -1.82\% & -0.72\% & -0.85\% & -1.40\% & -1.15\% & -1.44\% \bigstrut\\
\cline{2-8}      & BQTerrace & -2.57\% & -0.55\% & -0.70\% & -2.39\% & -0.55\% & -0.53\% \bigstrut\\
\cline{2-8}      & Class B Average & -2.39\% & -1.10\% & -1.07\% & -2.01\% & -1.13\% & -1.02\% \bigstrut\\
\hline
\multicolumn{1}{c|}{\multirow{5}[10]{*}{Class C}} & BasketballDrill & -2.12\% & -0.05\% & -1.45\% & -1.56\% & -0.35\% & -1.50\% \bigstrut\\
\cline{2-8}      & BQMall & -3.09\% & -1.02\% & -0.63\% & -3.00\% & -1.41\% & -0.02\% \bigstrut\\
\cline{2-8}      & PartyScene & -2.57\% & -2.18\% & -1.90\% & -2.48\% & -2.16\% & -2.21\% \bigstrut\\
\cline{2-8}      & RaceHorses & -1.44\% & -1.02\% & 0.21\% & -2.28\% & -1.85\% & -0.97\% \bigstrut\\
\cline{2-8}      & Class C Average & -2.31\% & -1.07\% & -0.94\% & -2.33\% & -1.44\% & -1.17\% \bigstrut\\
\hline
\multicolumn{1}{c|}{\multirow{5}[10]{*}{Class D}} & BasketballPass & -1.83\% & 0.22\% & -0.63\% & -2.08\% & -1.89\% & -0.52\% \bigstrut\\
\cline{2-8}      & BQSquare & -2.03\% & -2.17\% & 0.56\% & -2.06\% & -0.58\% & 1.81\% \bigstrut\\
\cline{2-8}      & BlowingBubbles & -2.83\% & -1.92\% & -0.39\% & -2.70\% & -2.16\% & 0.48\% \bigstrut\\
\cline{2-8}      & RaceHorses & -3.48\% & -2.99\% & -0.93\% & -3.52\% & -3.26\% & -1.83\% \bigstrut\\
\cline{2-8}      & Class D Average & -2.54\% & -1.71\% & -0.35\% & -2.59\% & -1.97\% & -0.01\% \bigstrut\\
\hline
\multicolumn{1}{c|}{\multirow{4}[8]{*}{Class E}} & Johnney & -3.82\% & -0.89\% & 0.43\% & -3.56\% & -1.33\% & -2.43\% \bigstrut\\
\cline{2-8}      & FourPeople & -4.03\% & -3.53\% & -4.22\% & -3.75\% & -3.58\% & -3.89\% \bigstrut\\
\cline{2-8}      & KristenAndSara & -3.19\% & -3.41\% & -1.00\% & -3.09\% & -3.52\% & -1.17\% \bigstrut\\
\cline{2-8}      & Class E Average & -3.68\% & -2.61\% & -1.60\% & -3.47\% & -2.81\% & -2.49\% \bigstrut\\
\hline
\multicolumn{2}{c|}{Average} & -2.65\% & -1.53\% & -0.96\% & -2.51\% & -1.73\% & -1.08\% \bigstrut\\
\hline
\end{tabular}%
}}
\end{table*}%

In real conditions, variable-block-size coding is used to \YY{further save bit-rate.} If the predictions for large blocks are accurate, there is no need to split the blocks and do separate prediction for \YY{each of the smaller blocks}. For HEVC, simply allowing a two-level split in block size brings 7\% bit-rate reduction. To evaluate our proposed model in real coding conditions, we remove the restriction on block size. The testing experiment is conducted on Common Test Conditions \cite{bossen2011common}. In our experiments, we set the scale of Coding Units (CU) to be up to $32\times32$, allowing split for intra prediction. The size of PU ranges from $4\times4$ to $32\times32$ and is adaptively \YY{decided by RDO. We train a different model for each coding block size, respectively.} As we evaluate the method in all intra configuration, only the first frame of each sequence is tested. The results are illustrated in gain in BD-Rate shown in TABLE \ref{table:detail}. The quantitative results show the improvement in rate-distortion performance using our proposed intra prediction method.
\RRR{
In addition to the solution with multiple models for variable block size, we also design a unified network PS-RNN+ to handle variable-block-size input. The basic units in PS-RNN+ are identical to those in the original PS-RNN, while only one set of the core PS-RNN units is used for all the block sizes.
The model of PS-RNN+ consists of three parts: pre-processing network (two stride convolutional layers), base network and post-processing network (two deconvolutional layers).} The base network is a PS-RNN trained with $8\times 8$ blocks as its input. When the block size is larger than $8\times 8$, preprocessing network is used for rescaling. We take an example where the block size is $32\times 32$.The input block is first down-sampled to $8\times 8$ by pre-processing network. Then, this block is fed into the base network, whose weights are shared to different block size input. After the prediction, the output of the last PS-RNN unit of the base network is up-sampled to the original size $32\times 32$ by post-processing network. To train the pre-processing and post-processing networks, we fix the weights of the base network and fine-tune those of others. Note that, the additional pre-processing and post-processing networks only introduce on average 6.6\% parameter increase. The network for the block-size of $4\times 4$ is kept as original, because up-sampling $4\times 4$ blocks will significantly increase the overall complexity without any performance improvement. The evaluation of BD-Rate for the unified integration PS-RNN+ is shown TABLE \ref{table:detail}, together with PS-RNN models. \RRR{As we can see from results, different implementation of PS-RNN models shows similar RD performance.}

\subsection{Comparison with Other Methods}
\RRR{We also compare the results with the method proposed in \cite{li2018tip} and \cite{dumas2018context}, which are both neural-network based methods for intra prediction, where variable-block-size configurations are also supported. The results are shown in TABLE \ref{tab:TIP_COMP}. For each pair of comparison, we use the same setting as described in the corresponding papers. The results for IPFCN are tested on four QPs (namely 22, 27, 32, 37) while the results for PNNS are tested on six QPs (namely 17, 22, 27, 32, 37, 42). We provide the results for these two settings, respectively. It can be seen from the results that the proposed method achieves a better RD performance compared with IPFCN-D, while it is comparable to PNNS. Different from PNNS, we dig for the potential of Spatial RNNs for intra prediction, which is specifically designed considering the progressiveness properly in this specific problem. The proposed method provides some additional insights for deep learning based intra predictor design.
\begin{table*}[htbp]

\captionsetup{labelfont={color=black},font={color=black}}
\centering
\caption{\RRR{Comparison on BD-Rate of PS-RNN with IPFCN-D \cite{li2018tip} and PNNS \cite{dumas2018context}. In the setting PS-RNN and IPFCN-D, the RD curve is interpolated from the results of four QPs (namely 22, 27, 32, 37), in PS-RNN Full and PNNS Full, the RD curve is interpolated from six QPs (namely 17, 22, 27, 32, 37, 42).}}
\label{tab:TIP_COMP}
\scalebox{1.1}{
\RRR{
\begin{tabular}{c|c|c|c|c|c}
\hline
Class & Sequence & PS-RNN  & PS-RNN Full & IPFCN-D \cite{li2018tip} & PNNS Full \cite{dumas2018context} \bigstrut\\
\hline
\multicolumn{1}{c|}{\multirow{5}[10]{*}{Class C}} & BasketballDrill & -1.6\% & -1.8\% & -1.5\% & \textbf{-3.5\%} \bigstrut\\
\cline{2-6}      & BQMall & -3.0\% & \textbf{-3.3\%} & -2.2\% & -3.1\% \bigstrut\\
\cline{2-6}      & PartyScene & -2.5\% & \textbf{-2.8\%} & -1.6\% & -2.4\% \bigstrut\\
\cline{2-6}      & RaceHorses & -2.3\% & -2.3\% & -3.2\% & \textbf{-3.3\%} \bigstrut\\
\hline
\multicolumn{1}{c|}{\multirow{5}[10]{*}{Class D}} & BasketballPass & -2.1\% & \textbf{-2.2\%} & -1.2\% & \textbf{-2.2\%} \bigstrut\\
\cline{2-6}      & BQSquare & -2.1\% & -2.4\% & -0.9\% & \textbf{-3.1\%} \bigstrut\\
\cline{2-6}      & BlowingBubbles & -2.7\% & \textbf{-2.7\%} & -1.9\% & \textbf{-2.7\%} \bigstrut\\
\cline{2-6}      & RaceHorses & -3.5\% & \textbf{-3.3\%} & -3.2\% & \textbf{-3.3\%} \bigstrut\\
\hline
\end{tabular}%
}
}

\end{table*}
}

\RRR{
Additionally, we make a comparison with the work in \cite{birman2018intra}, which utilize a fully connected neural network for intra prediction. As the method proposed in \cite{birman2018intra} is a multi-mode based one, we conduct this experiment on three typical modes. Following the configuration in \cite{birman2018intra}, the network predicts one pixel at a time. The results of MSE corresponding to the predicted pixel is shown in Table IV. The performance is tested in the Common Test Conditions. As shown in comparison, the proposed PS-RNN architecture results in lower MSE value for the predicted pixel for all modes. The experimental results show that PS-RNN produces better prediction results due to its better modeling spatial progressiveness in intra prediction than the FC network in \cite{birman2018intra}.

\begin{table}[htbp]
  \centering
  \caption{\RRR{MSE evaluation of the predicted pixel by PS-RNN and FC Net \cite{birman2018intra} respectively. Typical modes for common codecs, namely modes 7, 8 and 12 are tested.}}
    \begin{tabular}{c|ccc}
    \hline
    Methods & Mode 7 & Mode 8 & Mode 12 \bigstrut\\
    \hline
    PS-RNN & 114.44 & 116.65 & 133.82 \bigstrut[t]\\
    FC Net \cite{birman2018intra} & 130.38 & 126.34 & 148.97 \bigstrut[b]\\
    \hline
    \end{tabular}%

    \label{tab:modes}
\end{table}%

}

\begin{figure}[htbp]
    \centering
    \begin{subfigure}[b]{1.0\linewidth}
        \centering
    \includegraphics[width=1\linewidth]{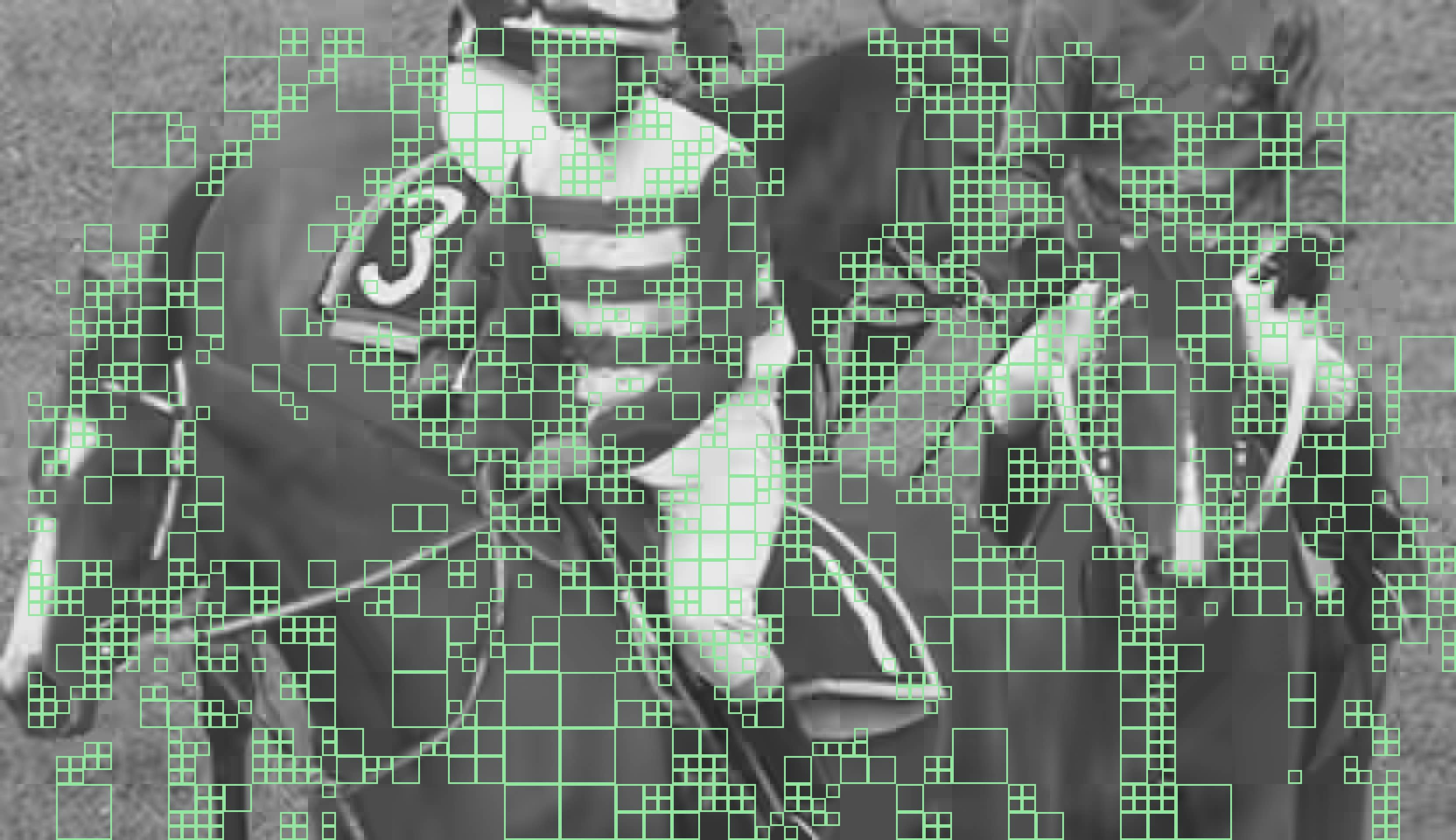}
    \caption{\RRR{For this frame in \textit{RaceHorses}, about 47\% of blocks are predicted using PS-RNN.}}
    \label{fig:racehorses}
  	\end{subfigure}
  	\begin{subfigure}[b]{1.0\linewidth}
        \centering
    \includegraphics[width=1\linewidth]{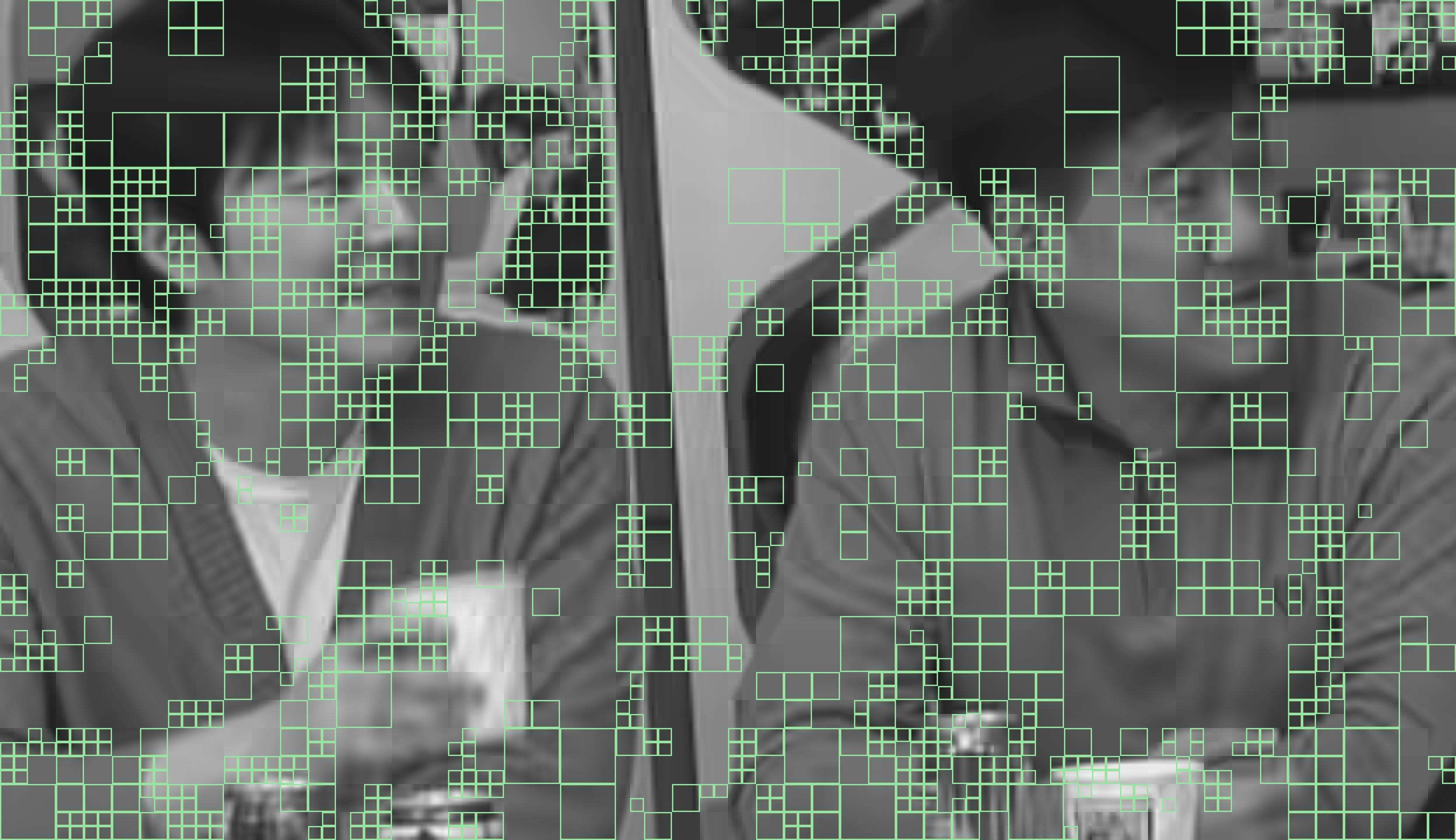}
    \caption{\RRR{We crop a region of size $416\times240$ from a frame in \textit{FourPeople}. For this frame, about 50\% of blocks are predicted using PS-RNN.}}
    \label{fig:party}
  	\end{subfigure}
    
    \caption{\RRR{Visual results of RDO selection. Blocks with green borders are predicted by PS-RNN.}}

	\label{fig:vis_frame} 
\end{figure}

We illustrate the results of RDO selection in \YY{Fig. \ref{fig:vis_frame}. On average about 35\% PUs are predicted using PS-RNN in all-intra mode}, which reflects the gain in performance.

\RRR{
\subsection{Robustness to Noise}
The network can better handle noise in the context compared with HEVC, as the network takes block-level pixels as the reference and the pixels are filtered by learned convolutional layers. As illustrated in Fig. \ref{fig:complex} (d), the directional prediction scheme in HEVC is interfered by the noise in the context. In Fig. \ref{fig:complex} I-(d), the results produced by HEVC are blurry as directional modes do not perform well in such non-smooth contexts. In Fig. \ref{fig:complex} III-(d), HEVC produces faulty lines. In TABLE \ref{tab:noise}, we compare the reduction in BD-Rate with different QPs. In high QP settings, the interference of the quantization noise has a large effect. As shown in TABLE \ref{tab:noise}, the network has a more significant reduction in performance for high QP settings, which shows its superior robustness to quantization noise.
}

\begin{table}[htbp]
\captionsetup{labelfont={color=black},font={color=black}}
\centering
\caption{\RRR{Evaluation of PS-RNN model on BD-Rate with settings of normal QPs \{22, 27, 32, 37\} and high QPs \{33, 38, 43, 48\}}.}
\label{tab:noise}

\RR{
\begin{tabular}{c|c|c}
\hline
QP Condition & Normal QP  & High QP  \bigstrut[t]\\
\hline
Class C & -2.59\% & -2.86\% \\
Class D & -2.33\% & -2.38\% \\
\hline
Average & -2.46\% & -2.62\% \bigstrut[b]\\
\hline
\end{tabular}%
}

\end{table}

\RRR{
\subsection{Complexity Analysis}
We evaluate the decoding time of the proposed network, and compare the results with the model in \cite{li2018tip}, where FC networks are utilized. The results are shown in TABLE \ref{tab:complexity}. Note that in the implementation, we use TensorFlow \cite{tensorflow2015-whitepaper} 1.12 as the backbone library and the binary is compiled with AVX2 support. In this paper we mainly focus on the analysis of the potential rate-distortion performance of PS-RNN-powered intra prediction scheme. To facilitate real world applications, acceleration of these methods is needed in future work.
}
\begin{table}[htbp]
\captionsetup{labelfont={color=black},font={color=black}}
\centering
\caption{\RRR{Decoding time of the PS-RNN network in HEVC (HM 16.15) and IPFCN~\cite{li2018tip}. HEVC (HM 16.15) is used as the anchor in the comparison.}}
\label{tab:complexity}
\RR{
\begin{tabular}{c|c|c}
\hline
Methods & PS-RNN & IPFCN \cite{li2018tip} \bigstrut\\
\hline
Decoding Time & 20664\% & 23011\% \bigstrut[t]\\
\hline
Class C BD-Rate & -2.30\% & -2.10\% \\
Class D BD-Rate & -2.60\% & -1.80\% \bigstrut[b]\\
\hline
\end{tabular}%
}
\end{table}

\section{Conclusion}\label{conclusion}
 In this paper we propose PS-RNN to improve intra prediction performance in video coding. \YY{The model is end-to-end trained to handle complex textures, making it capable of predicting for various content. The progressive generation of predictions addresses the problem of asymmetry in intra predition.} To further enhance the usability of the network in the codec, we propose to use SATD loss function for network training. It calculates both the distortion and the corresponding bit-rate for encoding the residue. \YY{Our network supports variable block size in HEVC}, bringing a leap in real coding performance. The proposed method achieves 2.5\% bit-rate reduction on average under the same reconstruction quality compared with HEVC.

\RR{
\appendix[Derivation of the Partial Derivatives of SATD Loss Function]

The network is trained using the back-propagation method. To train the network, we derive the partial derivative of the loss $S$ with respect to each entry of the distance $\mathbf{D}$. The formulation is presented as follows,
\begin{equation}
\begin{split}
\frac{\partial S}{\partial \mathbf{D}_{kl}} &= \frac{\partial\left(\sum_i\sum_j|\mathbf{D}'_{ij}|\right)}{\partial \mathbf{D}_{kl}}. \\ \newline
\end{split}
\end{equation}
Note that the absolute value function $y = |x|$ has no valid derivative at the point $x=0$, so we smooth the function by introducing a minor smoothing term $\epsilon$. Then the partial derivative turns to,

\begin{equation}
\begin{split}
\frac{\partial S}{\partial \mathbf{D}_{kl}} &=\frac{\partial\left(\sum_i\sum_j|\mathbf{D'}_{ij}|\right)}{\partial \mathbf{D}_{kl}} \\
&\approx \sum_i\sum_j\frac{\partial\sqrt{ {\mathbf{D'}_{ij}^2+\epsilon} }}{\partial \mathbf{D}_{kl}} \\
&=\sum_i\sum_j\frac{\partial\sqrt{ {\mathbf{D}_{ij}'^2+\epsilon}}}{\partial \mathbf{D}'_{ij}}\cdot\frac{\partial \mathbf{D'}_{ij}}{\partial \mathbf{D}_{kl}}\\
&=\sum_i\sum_j\frac{\mathbf{D'}_{ij}}{\sqrt{\mathbf{D'}_{ij}^2+\epsilon}}\cdot \frac{\partial \left(\mathbf{HDH}\right)_{ij}}{\partial \mathbf{D}_{kl}}\\
&=\sum_i\sum_j\frac{\mathbf{D'}_{ij}}{\sqrt{\mathbf{D'}_{ij}^2+\epsilon}}\cdot \frac{\partial \left(\mathbf{HD}\right)_{i\cdot}\mathbf{H}_{\cdot j}}{\partial \mathbf{D}_{kl}}.\\
\end{split}
\end{equation}
As only the terms involving $\mathbf{D}_{kl}$ in $\left(\mathbf{HD}\right)_{i\cdot}$ contribute to the partial derivative w.r.t. $\mathbf{D}_{kl}$, we remove other terms for simplification. We have
\begin{equation}
\begin{split}
\frac{\partial\left(\mathbf{HD}\right)_{i\cdot}\mathbf{H}_{\cdot j}}{\partial \mathbf{D}_{kl}} = \frac{\partial\left(\mathbf{H}_{ik}\mathbf{D}_{kl}\right)\mathbf{H}_{lj}}{\partial \mathbf{D}_{kl}},
\end{split}
\end{equation}
which implies that,
\begin{equation}
\begin{split}
\frac{\partial S}{\partial \mathbf{D}_{kl}} & \approx \sum_i\sum_j \frac{\mathbf{D'}_{ij}}{\sqrt{\mathbf{D'}_{ij}^2+\epsilon}}\cdot H_{ik}\cdot H_{lj}\\
&=\sum_i\sum_j\frac{\mathbf{D'}_{ij}}{\sqrt{\mathbf{D'}_{ij}^2+\epsilon}}\cdot\left(H_{ik}\cdot H_{jl}\right).
\end{split}
\end{equation}

}

{\footnotesize
\bibliographystyle{IEEEtran}
\bibliography{refs}
}

\begin{IEEEbiography}[{\includegraphics[width=1in,height=1.25in,clip,keepaspectratio]{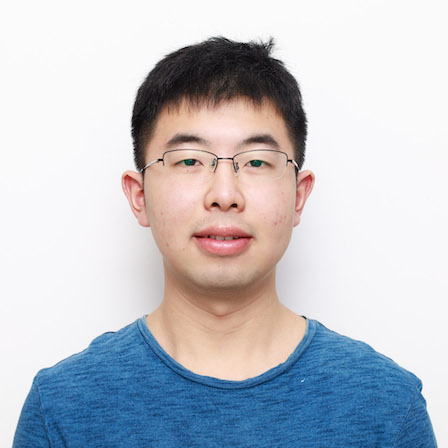}}]
	{Yueyu Hu}
	received the B.S. degree in computer science from Peking University, Beijing, China, in 2018, where he is currently pursuing the master’s degree with the Institute of Computer Science and Technology, Peking University. His current research interests include video and image compression and enhancement with machine learning.
\end{IEEEbiography}

\begin{IEEEbiography}[{\includegraphics[width=1in,height=1.25in,clip,keepaspectratio]{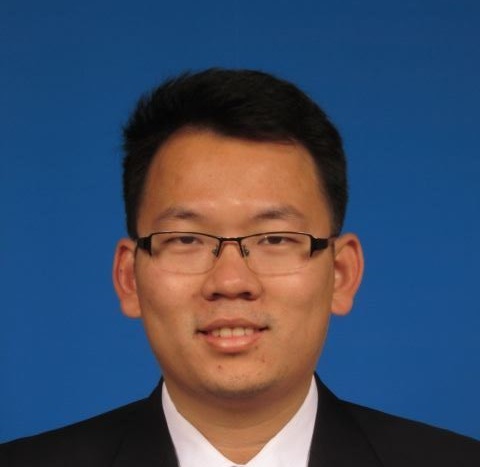}}]
	{Wenhan Yang} (17'S)
	received the B.S degree and Ph.D. degree (Hons.) in computer science from Peking University, Beijing, China, in 2012 and 2018.
	He is currently a postdoctoral research fellow with the Department of Computer Science, City University of Hong Kong, China. 
	Dr. Yang was a Visiting Scholar with the National University of Singapore, from 2015 to 2016. His current research interests include deep-learning based image processing, bad weather restoration, related applications and theories.
\end{IEEEbiography}

\begin{IEEEbiography}[{\includegraphics[width=1in,height=1.25in,clip,keepaspectratio]{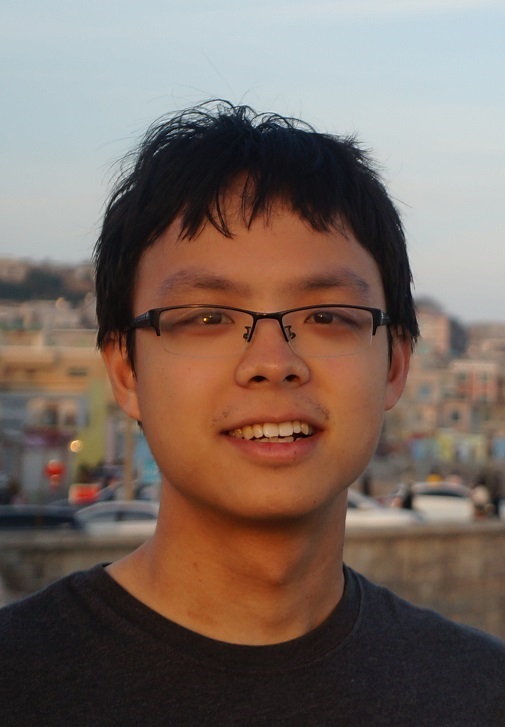}}]
		{Mading Li}
		received the B.S. degree in computer science from Peking University in 2013, and Ph.D. degree from Institute of Computer Science \& Technology of Peking University in 2018. He was a Visiting Scholar with McMaster University, ON, Canada from 2016 to 2017. He is currently an algorithm engineer with Video Technology Team at Kuaishou, China. He is currently focusing on image/video quality evaluation and smart video editing.
\end{IEEEbiography}

\begin{IEEEbiography}[{\includegraphics[width=1in,height=1.25in,clip,keepaspectratio]{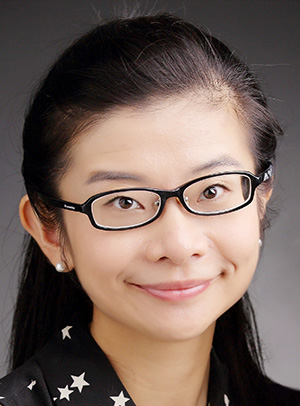}}]
	{Jiaying Liu} (S'08-M'10-SM'17)	
	is currently an Associate Professor with the Institute of Computer Science and Technology, Peking University. She received the Ph.D. degree (Hons.) in computer science from Peking University, Beijing China, 2010. She has authored over 100 technical articles in refereed journals and proceedings, and holds 34 granted patents. Her current research interests include multimedia signal processing, compression, and computer vision.

	Dr. Liu is a Senior Member of IEEE and CCF. She was a Visiting Scholar with the University of Southern California, Los Angeles, from 2007 to 2008. She was a Visiting Researcher with the Microsoft Research Asia in 2015 supported by the Star Track Young Faculties Award. She has served as a member of Multimedia Systems \& Applications Technical Committee (MSA-TC), Visual Signal Processing and Communications Technical Committee (VSPC) and Education and Outreach Technical Committee (EO-TC) in IEEE Circuits and Systems Society, a member of the Image, Video, and Multimedia (IVM) Technical Committee in APSIPA. She has also served as the Technical Program Chair of IEEE VCIP-2019/ACM ICMR-2021, the Publicity Chair of IEEE ICIP-2019/VCIP-2018, the Grand Challenge Chair of IEEE ICME-2019, and the Area Chair of ICCV-2019. She was the APSIPA Distinguished Lecturer (2016-2017). 

	In addition, Dr. Liu also devotes herself to teaching. She has run MOOC Programming Courses via Coursera/edX/ChineseMOOCs, which have been enrolled by more than 60 thousand students. She is also the organizer of the first Chinese MOOC Specialization in Computer Science. She is the youngest recipient of Peking University Outstanding Teaching Award.
\end{IEEEbiography}

\end{document}